\documentclass[runningheads]{llncs}

\usepackage[mobile]{eccv}

\usepackage{eccvabbrv}

\usepackage{graphicx}
\usepackage{booktabs}
\usepackage{times}
\usepackage{epsfig}
\usepackage{graphicx}
\usepackage{amsmath}
\usepackage{amssymb}
\usepackage{threeparttable}
\usepackage{booktabs}
\usepackage{caption}
\usepackage{colortbl}
\usepackage{rotating} % For sideways text
\usepackage{makecell}
\usepackage{float}
\usepackage{paralist}
\usepackage{xcolor}
\usepackage{threeparttable}
\usepackage{wrapfig,lipsum,booktabs}

\makeatletter
\newcommand\tabcaption{\def\@captype{table}\caption}
\makeatother
\usepackage[accsupp]{axessibility}  % Improves PDF readability for those with disabilities.

\usepackage[pagebackref,breaklinks,colorlinks,citecolor=eccvblue]{hyperref}
\begin{document}

%%%%%%%%% TITLE
\title{Intrinsic Image Decomposition Using Point Cloud Representation}
\author{Xiaoyan Xing\inst{1}, Konrad Groh\inst{2}, Sezer Karaoglu\inst{1},  Theo Gevers\inst{1}}
\institute{UvA-Bosch Delta Lab, University of Amsterdam \and Bosch Center for Artificial Intelligence, Robert Bosch GmbH}
\authorrunning{X. Xing et al.}
\maketitle

%%%%%%%%% ABSTRACT
\begin{abstract}

The purpose of intrinsic decomposition is to separate an image into its albedo (reflective properties) and shading components (illumination properties). This is challenging because it's an ill-posed problem. Conventional approaches primarily concentrate on 2D imagery and fail to fully exploit the capabilities of 3D data representation. 3D point clouds offer a more comprehensive format for representing scenes, as they combine geometric and color information effectively. 

To this end, in this paper, we introduce Point Intrinsic Net (PoInt-Net), which leverages 3D point cloud data to concurrently estimate albedo and shading maps. The merits of PoInt-Net include the following aspects. First, the model is efficient, achieving consistent performance across point clouds of any size with training only required on small-scale point clouds. Second, it exhibits remarkable robustness; even when trained exclusively on datasets comprising individual objects, PoInt-Net demonstrates strong generalization to unseen objects and scenes. Third, it delivers superior accuracy over conventional 2D approaches, demonstrating enhanced performance across various metrics on different datasets. \href{https://github.com/xyxingx/PoInt-Net}{Code}

% Intrinsic decomposition is a fundamental problem in computer vision. Traditionally, intrinsic decomposition methods mainly focus on 2D representations (i.e. images). In contrast to existing methods, in this paper, a 3D representation (point cloud) is used to approach the task of intrinsic decomposition.

% Our proposed method, Point Intrinsic Net, in short, \textbf{PoInt-Net}, jointly predicts the albedo, surface light direction, and shading. Our model provides light position relighting as a side application. 

% Large scale experiments are conducted on different datasets showing that, in most the proposed 3D representation method outperforms 2D representation approaches both quantitatively and qualitatively. Moreover, our method demonstrates good generalization for unseen objects and scenes. 
% Code \& Model: \url{https://anonymous.4open.science/r/PoInt-Net-ICCV-2847/}
\end{abstract}
%%%%%%%%% BODY TEXT
\section{Introduction}

% 1. SOTAs use multiple branch to predict geometry, heavy and rely on the dataset. -> off the shelf models do the dirty work,
% 2. Normal way to combine geometry information, e.g. RGBD image, is not efficient. -> point cloud make it more efficient. 

The aim of intrinsic decomposition is to separate an image into its albedo (reflective properties) and shading components (illumination properties). Decomposing an image into these fundamental components is a challenging problem due to its ill-posed nature, requiring specific constraints. Geometric data, including depth and surface normals, are used in facilitating this process \cite{garces2022survey}. Typically, these geometric cues are estimated as part of the process of intrinsic decomposition \cite{barron2015shape,baslamisli2021shadingnet,janner2017self,luo2020niid,sengupta2018sfsnet,zhu2022irisformer,li2021openrooms}. However, the success of this approach heavily depends on the accuracy of surface normal estimation and is sensitive to the 2D nature of the data employed, indicating a lack of adaptability across various data types. Traditional methods mainly focus on 2D images, overlooking the full potential of 3D data representation. 3D point clouds provide a richer scene representation by effectively integrating geometric and color information.

\begin{figure}[t]
    \centering
    \includegraphics[width=\textwidth]{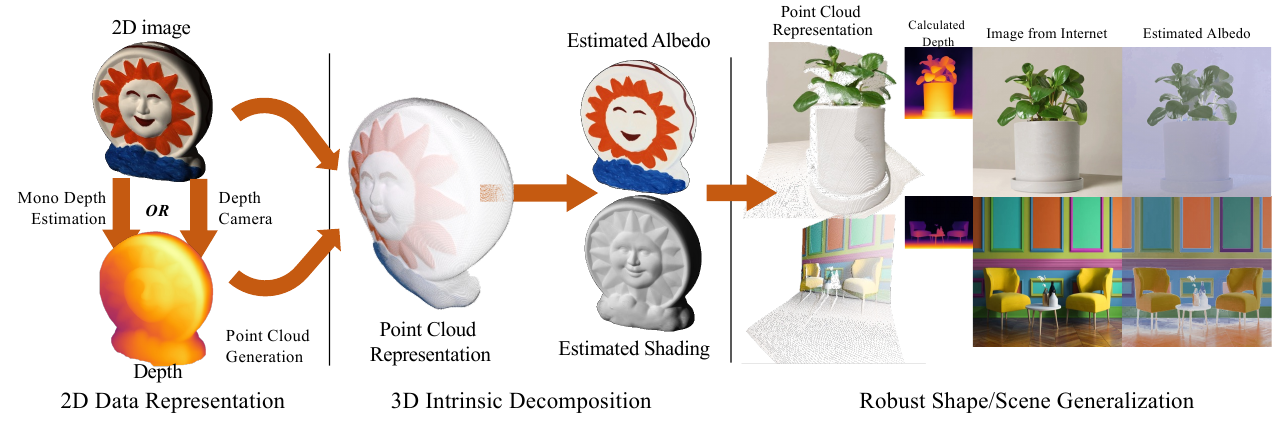}
    \caption{Intrinsic image decomposition using a 3D point cloud representation. Our approach decomposes the intrinsic components of an object/scene based on the point cloud representation of its appearance from a particular viewing angle. Point clouds are generated from \textit{RGB-D} images, where the depth maps are obtained by a depth camera (e.g. Lidar or ToF) or are estimated by a monocular depth estimation method such as \cite{Ranftl2022}. }
    \label{fig:teaser}
\end{figure}

This paper delves into utilizing 3D point clouds for the purpose of intrinsic decomposition. We focus on 3D point clouds obtained (1) directly from \textit{RGB-D} cameras, or (2) from 2D \textit{RGB} images where the depth $D$ map is measured by a monocular depth estimation technique, see Figure \ref{fig:teaser}. In this way, the proposed point cloud-based network, PoInt-Net, utilizes the 3D structure and appearance of objects or scenes to derive surface geometry and extract intrinsic features. Our method offers several benefits. First, point cloud representation naturally includes explicit 3D priors along with color details. Second, the intrinsic geometric information within the 3D point cloud is beneficial for more precise shading estimation, especially in areas where abrupt depth changes typically coincide with shifts in lighting \cite{xing2022point}. Third, point clouds accurately capture the shape of a scene, providing superior generalization for low-level vision tasks as shown by \cite{xu2022point}. Advancements in depth acquisition (such as Time-of-Flight cameras) and estimation technologies (for example, MiDaS \cite{Ranftl2022}) have substantially lowered the costs associated with acquiring depth information necessary for constructing point clouds.

Experimental analysis demonstrates that PoInt-Net excels in both efficiency and generalization capabilities. It outperforms existing models in shading estimation across multiple datasets with a reduced number of parameters, and still achieves impressive albedo outcomes. Trained exclusively on datasets containing singular objects, PoInt-Net showcases remarkable ability for zero-shot intrinsic estimation in real-world scenarios by using point clouds obtained from depth estimations.

The contributions of the paper are:
\begin{compactitem}
\item By applying intrinsic decomposition to a 3D point cloud framework, our approach innovatively merges geometric priors with sparse representations.
\item Introducing PoInt-Net, a point-based intrinsic image decomposition network with specialized subnets for light direction, shading, and albedo tasks.
\item PoInt-Net operates on sparse point clouds with far fewer parameters (1/10 to 1/100 of the existing methods), excelling on diverse datasets.
\item PoInt-Net facilitates zero-shot intrinsic estimation in real-world settings through the use of point clouds derived from estimated depths.
\end{compactitem}

\section{Related Work}

Intrinsic image decomposition (IID) is a complex, underdetermined problem that necessitates distinct constraints and priors, which can be sourced from images or supplementary data. We categorize prior IID approaches according to their input needs.
\paragraph{\textbf{IID with image(s).}}
One of the earliest work in 1970s \cite{barrow1978recovering} proposes to derive intrinsic characteristics from images. Numerous methods since then have employed simple network architectures for estimating these intrinsic characteristics \cite{cgintrinsic,narihira2015direct,shi2017learning,zhou2015learning,zoran2015learning,qian2021fast,fan2018revisiting}. \cite{buchsbaum1980spatial, grosse2009ground, baslamisli2018cnn, ma2018single} exploring IID by employing perceptual priors, based on the assumption that pronounced gradient edges denote changes in reflectance, while subtle edges suggest variations in illumination. \cite{shen2013intrinsic, rother2011recovering, shen2011intrinsic, das2022pie} present IID techniques that utilize clustering or reflectance sparsity. Exploring shading involves geometric cues, where \cite{barron2015shape} introduces a framework that capitalizes on geometric information extracted from images. \cite{baslamisli2021shadingnet,sengupta2018sfsnet,janner2017self,li2021openrooms,zhu2022irisformer} divide shading estimation into two processes (estimating surface normals and lighting) and then applying a shader for the final shading calculation. However, the success of these approaches depends on their precision in estimating surface normals, which complicates their adaptability to different types of data, e.g., \cite{sengupta2018sfsnet} is specifically designed for facial images and \cite{li2021openrooms,zhu2022irisformer} for indoor images. 
\paragraph{\textbf{IID with image(s) and additional input.}}
\textit{RGB-D} images are used for the purpose of IID. \cite{barron2015shape, chen2013simple, hachama2015intrinsic, kim2016unified} show that incorporating extra geometric cues (such as depth and surface normals) to refine the shading components results in enhanced decomposition outcomes. \cite{lee2012estimation} proposes a model for IID using a sequence of \textit{RGB-D} video frames. Recently, \cite{sato2023unsupervised} leverages the LiDAR intensity to separate the reflectance from an image. In addition to \textit{RGB-D} data type, \cite{laffont2012coherent} uses multi-view stereo to reconstruct 3D points and surface normals to estimate the intrinsic components. \cite{zhang2021nerfactor} employs a Neural Radiance Field (NeRF) \cite{mildenhall2020nerf} to obtain the intrinsic components in an implicit representation. \cite{ye2023intrinsicnerf} utilizes multi-view constraints and semantic labels to derive intrinsic properties from a NeRF. The inclusion of additional input aids models in understanding the geometric relationships between pixels. Nonetheless, the above approaches encounter two primary challenges: 1) it lacks computational efficiency for \textit{RGB-D} image-based decomposition because of the expensive 3D operations involved; 2) it suffers from limited generalization capabilities in multi-view-based approaches as the Neural Radiance Fields (NeRF) are prone to overfitting specific scenes.

In contrast to previous \textit{RGB-D} methods, our approach utilizes point cloud representation for estimating intrinsic components, without explicitly estimating surface normals. This strategy leads to enhanced robustness in decomposition, yielding a point-based operation network for intrinsic image decomposition, which improves accuracy and reduces computational costs.

\vspace{-1ex}

\section{Point Cloud Intrinsic Representation}
\vspace{-2ex}
In this section, we introduce a new intrinsic representation technique utilizing point clouds. Section~\ref{Sec:Render} details the intrinsic decomposition process grounded in the rendering model. Section~\ref{Sec:Problem} revisits the intrinsic decomposition from the perspective of a point cloud representation.
% Section~\ref{Sec:Networks} provides the technical details of the network architecture; and Section~\ref{Sec:Training} introduces the learning strategy to train the network.
\begin{figure*}[!ht]
    \centering
    \resizebox{\textwidth}{!}{\includegraphics{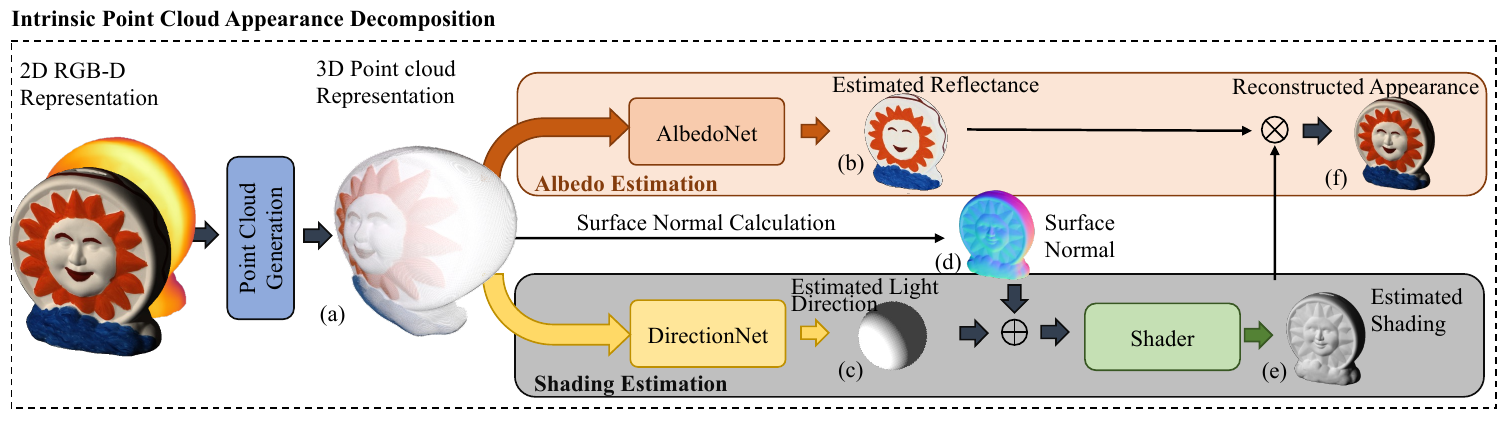}}
    \caption{Our proposed framework for intrinsic point cloud decomposition starts by transforming the \textit{RGB-D} representation into a point cloud representation. (a) The point cloud representation is used as input to train two separate components: the shading and the albedo estimations. The shading estimation is supported by the DirectionNet (Light Direction Estimation Net), which takes (a) as input and and outputs surface light direction estimates (c). Surface normals (d) are calculated using local neighborhoods within (a). The Shader (Learnable Shader) then uses the concatenated vectors of (c) and (d) to generate the final shading estimation (e). The albedo estimation is obtained by the AlbedoNet (Point-Albedo Net) which extracts invariant reflectance (b) from (a) based on the Lambertian assumption. Finally, by multiplying (b) and (e), the reconstructed image (f) is generated. Please refer to the supplementary for a more detailed explanation.}
    \label{Fig:main_framework}
\end{figure*}

\subsection{Intrinsic Decomposition}
\label{Sec:Render}
For a given point $x$, the reflected radiance $L$ is defined by~\cite{kajiya1986rendering}:
\begin{equation}
    L(x,\omega_o) = \int_{\omega_i \in \Omega+}f_r(x,\omega_i,\omega_o) L_i(x,\omega_i)(N \cdot \omega_i) d\omega_i,
\end{equation}
where $\omega_i$ is the light angle from the upper hemisphere $\Omega_+$, $\omega_o$ is the viewing angle, $N$ is the surface normal, $L_i (x,\omega_i )$ is the position of the lighting angle and its direction, and $f_r$ is the surface reflectance, modeled by a Bidirectional Reflectance Distribution Function (BRDF) \cite{nicodemus1965directional}.

Given a viewing angle, assuming that the surface is Lambertian, the diffuse appearance $I_{diffuse}$ is given by:
\begin{equation}
\label{Equ:Int_diffuse}
    \textbf{I}_{diffuse} =  \int_{\omega_i \in \Omega+}f_r(\omega_i)L_i(\omega_i)(N \cdot \omega_i) d\omega_i.
\end{equation}
Conventionally, $\frac{\rho_d}{\pi}$ denotes the reflectivity of the surface (albedo), where $f_r(\omega_i)=\frac{\rho_d}{2\pi}$. Therefore, if the illumination is uniform, the model is defined by:
\begin{equation}
    \textbf{I}_{diffuse} = \frac{\rho_d}{\pi} \cdot (\textbf{N} \cdot \textbf{L}_{in}), 
\end{equation}
where, $\textbf{L}_{in}$ represents the visible incident light. The aim of intrinsic decomposition is to disentangle the albedo $\textbf{A}=\frac{\rho_d}{\pi}$ and shading $\textbf{S}=(N\cdot L_{in})$ from the appearance $\textbf{I}_{diffuse}$, where $(\cdot)$ is the dot product. 

\subsection{Intrinsic Appearance of Point Clouds}
\label{Sec:Problem}

According to Equation \eqref{Equ:Int_diffuse}, the appearance of an object under a given lighting condition is represented by a \textit{RGB} image $\textbf{I} = [\textbf{I}_r,\textbf{I}_g,\textbf{I}_b] \in\mathbb{R}^{U\times{V}\times3}$. Additionally, its corresponding depth map is given by $\mathrm{D}\in\mathbb{R}^{U\times{V}\times1}$. Depth data can be acquired from a depth camera, such as LiDAR, or it can be estimated from 2D images using a monocular depth estimation method, e.g.,\cite{Ranftl2022}.

The \textit{RGB} image and its associated depth map are converted into a colored point cloud representation, $\textbf{P} = \{\textbf{p}_i | i \in 1, \ldots, n\}$. Each point $\textbf{p}_i$ is represented as a vector of $[x,y,d,r,g,b]$ values:
\begin{equation}
\label{equ:build pcd}
\textbf{p}_i = \left(\frac{(u-c_x)d}{f_x}, \frac{(v-c_y)d}{f_y}, d,r,g,b\right),
\end{equation}
where, $f_x$ and $f_y$ are the focal lengths, and $(c_x, c_y)$ is the principal point.

Given a dataset of $M$ point clouds, $\mathcal{P}=\{\textbf{P}_{1},\textbf{P}_{2},...,\textbf{P}_{M}\}$, its intrinsic components are defined by: 1)  Albedo $\mathcal{A}=\{\textbf{A}_{1},\textbf{A}_{2},...,\textbf{A}_{M}\}$, 2) Shading $\mathcal{S}=\{\textbf{S}_{1},\textbf{S}_{2},...,\textbf{S}_{M}\}$, 3) Surface normal $\mathcal{N}=\{\textbf{N}_{1},\textbf{N}_{2},...,\textbf{N}_{M}\}$, and 4) Light source position $\mathcal{L}=\{\textbf{L}_{1},\textbf{L}_{2},...,\textbf{L}_{M}\}$.

\textbf{Albedo}
contains the reflectance information. Hence, a direct point based mapping ($f_{\alpha}:\mathcal{P}\to\mathcal{A}$) is employed to decompose the reflectance from input point cloud.

\textbf{Shading}
depends on the object geometry, viewing and lighting conditions. Thus, instead of directly learning the shading, a point-light direction net ($f_{\theta}:\mathcal{P}\to\mathcal{L}$) is used to estimate the light direction from the point cloud representation. Then, a point-learnable shader ($f_{\sigma}:\mathcal{L},\mathcal{N}\to\mathcal{S}$) is trained to generate the rendering effects based on the surface normals (from input point cloud) and light direction estimation (from point-light direction net).
%benefit 

Adopting point cloud representation for intrinsic decomposition is beneficial due to several reasons: 1) It naturally integrates depth and \textit{RGB} information into a cohesive structure. 2) It facilitates the derivation of surface normal data, ascertainable through local neighborhood analysis. 3) It provides a resilient framework capable of withstanding errors in depth measurements, as our additional ablation study indicates that minor inaccuracies in certain points have a negligible impact on the overall representation.

\section{Point Based Intrinsic Decomposition}
In this section, a novel point based intrinsic decomposition technique is proposed. Section~\ref{Sec:Networks} provides the technical details of the network architecture and Section~\ref{Sec:Training} introduces the learning strategy to train the network.

\subsection{Point Intrinsic Net}
\label{Sec:Networks}
Our proposed network, PoInt-Net, consists of three key components: 1) the Point Albedo-Net, which is designed to capture the properties of surface materials, 2) the Light Direction Estimation Net, tasked with identifying lighting conditions to support the albedo estimation of the point cloud, and 3) the Learnable Shader, a module that combines the deduced light direction with surface normals to generate the shading map. The approach most similar to ours is \cite{janner2017self}. However, it needs ground-truth normal information to initialize and is limited by its generalization capabilities to individual objects.

Figure \ref{Fig:main_framework} shows the architecture of the proposed network and the specifics of its forward connections. The design of the three sub-networks is largely consistent, with minor variations such as the activation functions used. Specifically, all three sub-nets are adopted from \cite{qi2017pointnet}, and employ Multi-Layer Perceptrons (MLPs) for point-feature extraction and decoding, with the aim of solving the point-to-point relationship. 

\textbf{Point albedo-net} inputs a 6-D point cloud, which includes color data and spatial coordinates, and generates surface reflectance estimates. The Rectified Linear Unit (ReLU) is employed as the activation function to yield scaled output colors.

\textbf{Light direction estimation Net} takes the same input as \textit{Point Albedo-Net}, and predicts point-wise surface light directions. ReLU is used as the activation function in most of the layers. The final two layers use the hyperbolic tangent function (Tanh) to ensure that all light directions are estimated.

\textbf{Surface normal calculation} computes the surface normal from the given point cloud through several steps: 1) identifying neighboring points and calculating the covariance matrix, 2) computing the eigenvectors of the covariance matrix, and 3) selecting the normal vector associated with the smallest eigenvalue.
% To enhance the efficiency of the training phase, normal information is pre-calculated and utilized throughout training.

\textbf{Learnable shader} takes concatenated vectors of surface normal information (derived from the input point cloud) and surface light direction estimation as input, and produces a point-wise shading map as output.

Our choice of module configuration is based on empirical evidence:
1) \textit{Module Specialization.}: Given PointNet's limitations as an relatively weaker encoder, we use three specialized modules for separate estimation of albedo, lighting, and shading. Which also guarantees the robustness in the generalization. 
2) \textit{Input Integration.} We found that concatenating surface normals with lighting, rather than multiplying them, yields improved outcomes.

\subsection{Joint-learning Strategy}
\label{Sec:Training}
A two-step training strategy is employed to arrive at an end-to-end intrinsic decomposition learning pipeline. First, in terms of shading estimation, the \textit{Light Direction Estimation Net} and \textit{Learnable Shader} are trained using the ground-truth light position $\textbf{L}$ and shading map $\textbf{S}$. Then, for albedo estimation, the parameters in these two sub-nets are preserved and frozen, while the \textit{Point Albedo-Net} is constrained by the ground-truth albedo $\textbf{A}$ and the final reconstructed image $\hat{\textbf{I}}$ (multiplied by the estimated albedo map $\hat{\textbf{A}}$ and the estimated shading map $\hat{\textbf{S}}$). During training, the mean square error is used. The loss function\footnote{For datasets without light source labels, the loss function only contains the shading map $\hat{\textbf{S}}$.}, for stage one, is:
\begin{equation}
    \mathcal{L}_{shading} = \frac{1}{M}\sum^M({|\textbf{L}-\hat{\textbf{L}}|^2+|\textbf{S}-\hat{\textbf{S}}|^2}).
\end{equation}
For stage two, a number of loss functions are used. To address reflectance changes, a color cross ratio loss inspired by \cite{gevers1999color} is taken, formulated as follows:
\begin{equation}
    \mathcal{L}_{ccr} = |M_{RG}-M_{\hat{R}\hat{G}}|+|M_{RB}-M_{\hat{R}\hat{B}}| + |M_{GB}-M_{\hat{G}\hat{B}}|,
\end{equation}
where $\{M_{RG},M_{RB},M_{GB}\}$, $\{M_{\hat{R}\hat{G}},M_{\hat{R}\hat{B}},M_{\hat{G}\hat{B}}\}$ are the cross color ratios from the ground-truth albedo and the estimated albedo respectively. Please refer to supplemental for the details of the cross color ratios calculation. 
Similar to \cite{fan2018revisiting}, the gradient difference is considered and is formulated by:
\begin{equation}
    \mathcal{L}_{grad} = |\nabla \textbf{A}-\nabla \hat{\textbf{A}}|^2_2.
\end{equation}
Hence, the reconstruction loss is applied to constrain the estimated albedo:
\begin{equation}
    \mathcal{L}_{rec} = \frac{1}{M}\sum^M({|\textbf{A}-\hat{\textbf{A}}|^2+|\textbf{I}-\hat{\textbf{I}}|^2}).
\end{equation}
The final loss function is given by:
\begin{equation}
    \mathcal{L}_{albedo} = \mathcal{L}_{rec}+\mathcal{L}_{grad}+\mathcal{L}_{ccr},
\end{equation}
where $\{\hat{\cdot}\}$ represents the estimated values, and $M$ is the number of input point clouds in the mini-batch. Adam \cite{kingma2014adam} is employed as the optimizer. 

\section{Experiments} 
This section outlines the experimental framework used to assess the effectiveness of our proposed approach.

\paragraph{\textbf{Dataset}.}
Five publicly accessible datasets are employed in the assessment:
\begin{compactitem}
    \item \textit{ShapeNet-Intrinsic} \cite{janner2017self}: Based on  ShapeNet \cite{shapenet2015}, albedo and shading are generated by the Blender-cycle. The dataset contains ground-truth depth, normal, and light position information. We follow the same dataset split as Liu \etal \cite{liu2020cvpr}.
    \item \textit{MIT-Intrinsic} \cite{grosse2009ground}: For this real-world dataset, albedo and shading under different illumination conditions are provided. Depth information, calculated by \cite{barron2015shape}, is used. The train and test split are kept the same as proposed by \cite{barron2015shape}.
\item \textit{MPI-Sintel} \cite{butler2012naturalistic}: This synthetic dataset provides albedo, shading, and depth information. 
We use the same training and test splits as those employed by existing methods to assess our approach.
\item \textit{Inverender} \cite{zhang2022modeling}: This synthetic dataset includes ground-truth albedo and normals. To evaluate the performance of our method, we adhere to the same training and test splits established by previous methods \cite{ye2023intrinsicnerf, zhang2021nerfactor}.
\item \textit{IIW} \cite{bell2014intrinsic}: 
A real-world image dataset comprising a wide variety of scenes and lighting conditions, along with the weighted human disagreement rate (WHDR) label.
\end{compactitem}
Moreover, we employ various images sourced from the internet to demonstrate our capability to generalize to real-world scenarios.

\paragraph{\textbf{Depth acquisition.}} For depth information, we utilize ground-truth depth from the datasets where available. In cases without ground-truth depth, including certain datasets \cite{zhang2021nerfactor,bell2014intrinsic} and real-world images, we apply the mono-depth estimation method \cite{Ranftl2022} to generate relative depth information for point cloud construction. 
% This ensures consistent depth data handling across different sources.

\paragraph{\textbf{Metrics and visual comparison}} For quantitative evaluation, we employ three commonly used metrics: Mean Square Error (MSE), Local Mean Squared Error (LMSE), and Structural Dissimilarity (DSSIM). These metrics are our standard evaluation criteria, unless specified otherwise. For fair qualitative comparison, visualization images for all benchmarks are objectively selected, sourced directly from the publications of the methods being compared or acquired from their official implementations.

\paragraph{\textbf{Pretrained on ShapeNet-Intrinsic dataset.}} Due to the limited size of the intrinsic decomposition datasets, such as MIT-intrinsic \cite{grosse2009ground} with only 20 objects and MPI-Sintel \cite{butler2012naturalistic} with hundreds of images, employing pre-trained models is a common practice. For instance, \cite{qian2021fast, cheng2018intrinsic} use pre-trained weights from ImageNet \cite{krizhevsky2012imagenet}. Similarly, \cite{das2022pie} pre-train on the NED dataset \cite{baslamisli2018joint}, while \cite{janner2017self}, \cite{liu2020cvpr}, and \cite{baslamisli2018cnn} use ShapeNet for pre-training. 
In our comparative tables, we indicate which models have undergone pre-training and specify the datasets utilized for this pre-training. In this way, fairness and clarity is ensured in our evaluation.

% \paragraph{Training/Tuning settings.} Our method undergoes evaluation across various datasets. Unless explicitly stated otherwise, the term "pretrained on single object-level model" refers to the version trained on the ShapeNet-Intrinsic dataset \cite{janner2017self}, which includes ground-truth labels for intrinsic images and light positions.
\begin{figure}[!t]
    \centering
    \resizebox{\linewidth}{!}{\includegraphics{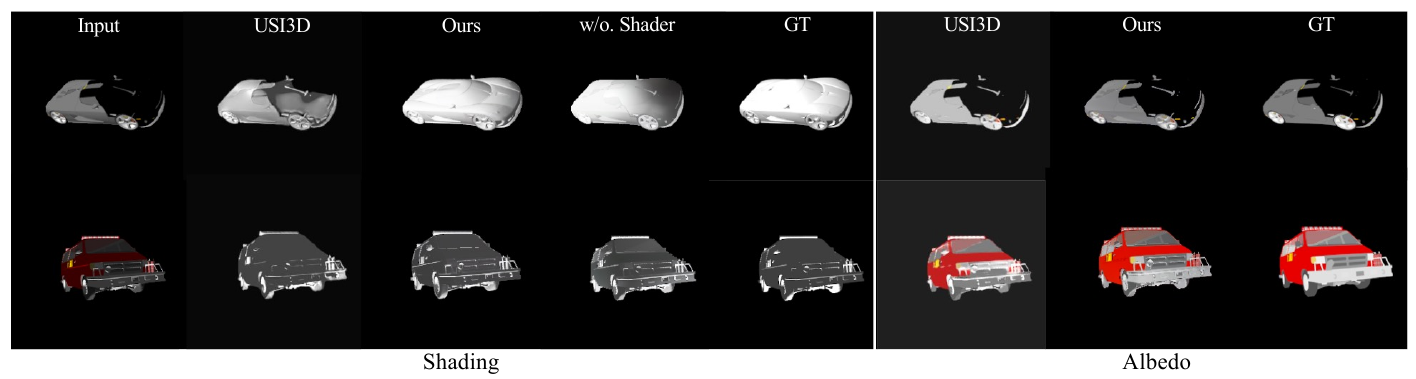}}
    \caption{Comparison to state-of-the-art method USI3D (fine-tuned version) \cite{liu2020cvpr} and ablation study on the ShapeNet-intrinsic dataset. \textbf{Zoom to see details.}}
    \label{Fig:SN-intrinsic}
\end{figure}

\subsection{Evaluation on IID datasets}
\label{Sec:Obj-result}

\paragraph{\textbf{ShapeNet-Intrinsic}}
\begin{wrapfigure}{l}{6cm}
\centering
        \resizebox{\linewidth}{!}{
        \begin{threeparttable}\begin{tabular}{lccc|c|c}
            \toprule[1.2pt]
            &\multicolumn{3}{c}{MSE$(10^{-2})$}&{LMSE$(10^{-2})$}&{DSSIM$(10^{-2})$} \cr\cline{2-6}
            &Albedo&Shading&Average&Total&Total\cr
            \hline
            \specialrule{0em}{1pt}{1pt}
            CGIntrinsics\cite{li2018learning}  	   &	3.38  &   2.96 &    3.17   &  6.23&- \\
            Fan \etal~\cite{fan2018revisiting}  &  3.02 &   3.15  & 	   3.09 &    7.17 &-\\
            Ma \etal~$*$\cite{ma2018single}     &  2.84 &  2.62  &   2.73 &    5.44 & -\\
            USI3D$*$ ~\cite{liu2020cvpr}  & {1.85} &  {1.08}    & {1.47}  & {4.65} &-\\
            \midrule
            Ours (w/o. shader)  & \underline{0.48} & \underline{0.57}   & \underline{0.53} &  \underline{1.15} &{4.93}\\
            Ours  & \textbf{0.46} & \textbf{0.38}  &\textbf{0.42}   &\textbf{1.00} &{4.15}\\
            \bottomrule[1.2pt]
    \end{tabular}
    \begin{tablenotes}
        \footnotesize
        \item[$*$] Unsupervised methods but finetune on the dataset. 
    \end{tablenotes}
    \end{threeparttable}}
     \tabcaption{{Results and ablation study for ShapeNet-Intrinsic} \cite{janner2017self}. }
\label{Tab:ShapeNet}
\vspace{-3ex}
\end{wrapfigure}

Table \ref{Tab:ShapeNet} shows a quantitative comparison of PoInt-Net with the latest open source methods \cite{li2018learning,ma2018single,fan2018revisiting,liu2020cvpr}. Our results clearly indicate that our approach significantly surpasses existing methods across all three metrics. To provide specific values, PoInt-Net achieves an MSE of 0.0046 for albedo and 0.0038 for shading, with an LMSE of 1.00 and a DSSIM of 0.0415. This exceptional performance is attributed to our method's capacity to capture and utilize intricate relationships among intrinsic properties, leading to a more robust and dependable estimation.

Qualitative results of the method are shown in Figure \ref{Fig:SN-intrinsic}, where USI3D \cite{liu2020cvpr} (fine-tuned version) is used as a reference. PoInt-Net generates a shading map by utilizing surface light direction and normal information, effectively separating shading from the composite image. This process results in realistic and consistent shading in the output images, closely reflecting the underlying surface geometry. Particularly noticeable is the clear distinction between shading and surface in darker areas of objects, underscoring the method's efficacy and robustness in generating high-quality, visually appealing outputs that faithfully represent intrinsic properties of objects.

\begin{figure*}[!t]
    \centering
    \resizebox{\textwidth}{!}{\includegraphics{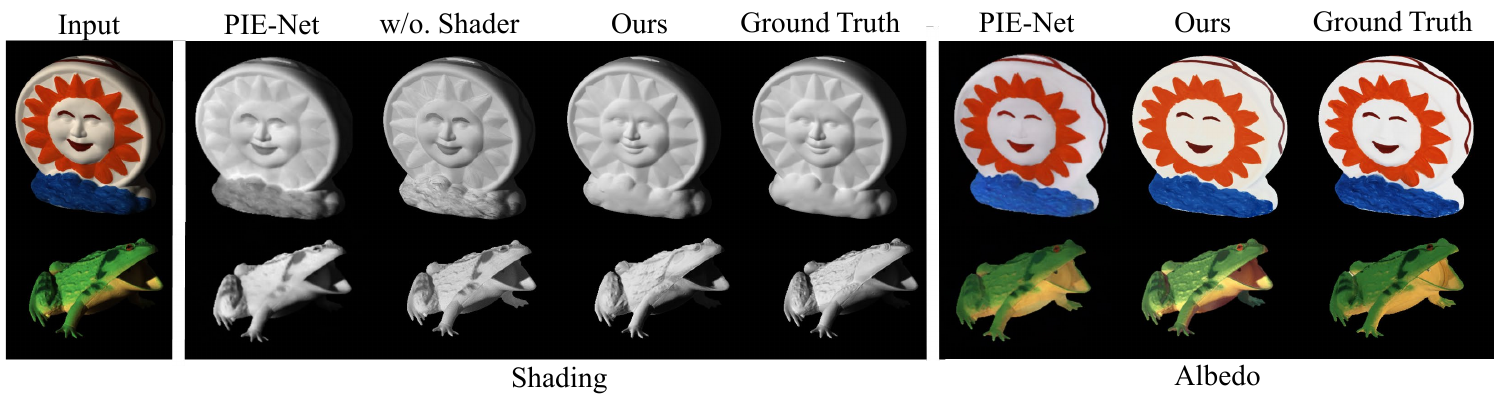}}
    \caption{Qualitative results on the MIT-intrinsic benchmark \cite{grosse2009ground}. Comparison to the state-of-the-art method PIE-Net~\cite{das2022pie}. Ablation study on shader is conducted.}
    \label{Fig:MIT-intrinsic}
\end{figure*}
\vspace{-2ex}
\begin{wrapfigure}{r}{7cm}
\renewcommand{\arraystretch}{1} % Default value: 1
\centering
\resizebox{0.8\linewidth}{!}{
\begin{threeparttable}\begin{tabular}{lcc|cc|cc}
% \vspace{-1ex}
\toprule
~ &
\multicolumn{2}{c}{MSE$(10^{-2})$}&\multicolumn{2}{c}{LMSE$(10^{-2})$}&\multicolumn{2}{c}{DSSIM$(10^{-2})$} \cr\cline{2-7}
& Albedo&Shading&Albedo&Shading&Albedo&Shading\cr
\midrule
SIRFS \cite{barron2015shape}&  1.47 & 1.83 & 4.16 & \underline{1.68}& 12.38&9.85\cr
Zhou \textit{et al.} \cite{zhou2015learning} & 2.52 & 2.29  & -& -&-&-\cr
Shi \textit{et al.} \cite{shi2017learning} & 2.78 & 1.26&5.03  & 2.40&14.65&12.00 \cr
DI~\cite{narihira2015direct}&  2.77 & 1.54& 5.86&2.95&15.26&13.28 \cr

Ma \etal$*$ \cite{ma2018single}&  3.13 & 2.07  & 1.16& 0.95&-&-\cr

Janner~\etal$*$\cite{janner2017self}& 3.36 & 1.95  & 2.10 &1.03&-&-  \cr
CGIntrinsics~\cite{li2018learning}&  1.67 & 1.27  & 3.19&2.21&12.87&13.76  \cr

USI3D$*$\dag \cite{liu2020cvpr} ~ & 1.57 &  1.35  &  1.46 &2.31&-&-\cr 
FFI-Net\dag \cite{qian2021fast}  ~   &1.11 &  0.93 & 2.91 &  3.19&10.14&11.39 \cr % 0.1255 & 0.1129
PIE-Net\dag \cite{das2022pie} ~ & \textbf{0.28} &  \underline{0.35}  & \underline{1.36}& 1.83& \textbf{3.40} &\underline{4.93}\cr 
\midrule

Ours\dag ~ & \underline{0.89} &  \textbf{0.34}  &  \textbf{0.97}&\textbf{0.37}& \underline{4.39}&\textbf{3.02} \cr

\bottomrule
\end{tabular}
\begin{tablenotes}
\footnotesize
        \item[$*$] Unsupervised methods but finetuned on the dataset. 
        \item[\dag] Using pre-trained parameters.
    \end{tablenotes}
\end{threeparttable}}
\tabcaption{
{Results for MIT Intrinsic}. 
}
\label{tab:mit}
\end{wrapfigure}

\paragraph{\textbf{MIT-intrinsic}}
In addition to the synthetic dataset, this section encompasses an evaluation on the MIT-intrinsic dataset to assess the proposed method's ability to generalize to real-world scenarios. The results obtained from the MIT-intrinsic dataset are consistent with those from the synthetic dataset, validating the method's efficacy and robustness across varied datasets.

Table \ref{tab:mit} reports the quantitative results. PoInt-Net produces state-of-the-art shading results on the MIT-intrinsic dataset for all metrics. Moreover, our approach obtains the best LMSE  and the second-best performance for the albedo output in terms of MSE and DSSIM metrics. Note that \cite{das2022pie} employs an extra input to provide additional information for albedo estimation.

The visualization results are given in Figure \ref{Fig:MIT-intrinsic}. Our method outperforms the latest state-of-the-art method \cite{das2022pie} in accurately distinguishing the markings on the frog's back. This success is due to PoInt-Net's incorporation of surface light direction estimation and surface normal calculation, which contribute to the production of high-quality shading outcomes.

\textbf{Note:} Ground-truth depth information is not included in the MIT-intrinsic dataset. To this end, depth information is used and computed by the method of \cite{barron2015shape}. Although depth estimations frequently contain noise, such as outliers and invalid points, PoInt-Net consistently learns intrinsic features even when the input data includes noise. This underscores its robustness to imperfect depth information.
% \footnote{Available for download at: \href{https://github.com/bhushan23/SIRFS/tree/master/data/MIT-Berkeley-Laboratory}{here}}

\setlength{\tabcolsep}{1.5pt}

\begin{figure*}[t]
    \centering
    \resizebox{\textwidth}{!}{\includegraphics{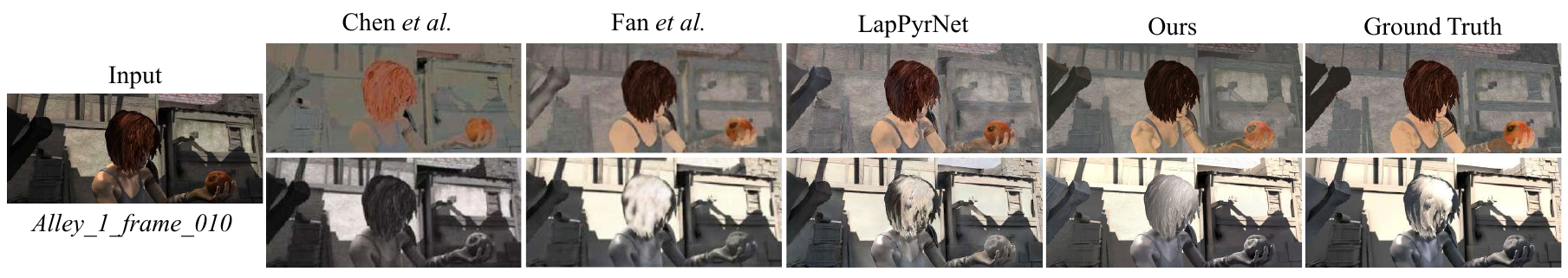}}
    \caption{Visual results (\textit{image split}) for the MPI Sintel dataset \cite{butler2012naturalistic}. Comparison to state-of-art-methods Fan \etal\cite{fan2018revisiting} and LapPyrNet\cite{cheng2018intrinsic}. The method of Chen \etal \cite{chen2013simple} is provided as reference, since it uses \textit{RGB-D} images as input.}
    \label{Fig:MPI-intrinsic}
\end{figure*}
\begin{table}[!t]
\centering
\resizebox{0.75\linewidth}{!}{
\small
\begin{threeparttable}\begin{tabular}{l |ccc | ccc |ccc}

\toprule[1.2pt]
&
\multicolumn{3}{c}{si-MSE$(10^{-2})$}&\multicolumn{3}{c}{si-LMSE$(10^{-2})$}&\multicolumn{3}{c}{DSSIM$(10^{-2})$}\\%\cr\cline{2-10}
&Albedo&Shading&Average&Albedo&Shading&Average&Albedo&Shading&Average\\
\midrule
Retinex \cite{grosse2009ground} ~&6.06&7.27&6.67&3.66&4.19&3.93&22.70&24.00&23.35\cr%\hline
Lee \etal \cite{lee2012estimation}~&4.63&5.07&4.85&2.24&1.92&2.08&19.90&17.70&18.80\cr%\hline
SIRFS ~ \cite{barron2015shape}&4.20&4.36&4.28&2.98&2.64&2.81&21.00&20.60&20.80\cr%\hline
Chen\etal~\cite{chen2013simple}&3.07&2.77&2.92&1.85&1.90&1.88&19.60&16.50&18.05\cr%\hline
DI~\cite{narihira2015direct}&1.00&0.92&0.96&0.83&0.85&0.84&20.14&15.05&17.60\cr%\hline
DARN~\cite{lettry2018darn}&1.24&1.28&1.26&0.69&0.70&0.70&12.63&12.13&12.38\cr%\hline
Kim\etal~\cite{kim2016unified}&0.70&0.90&0.70&0.60&0.70&0.70&\c9.20&10.10&9.70\cr%\hline
Fan\etal \dag ~\cite{fan2018revisiting}& 0.69& \textbf{0.59}& \underline{0.64}& \underline{0.44}& \underline{0.42}& \underline{0.43} &11.94&8.22&10.08\cr
LapPyrNet\dag~\cite{cheng2018intrinsic}& \underline{0.66} & \underline{0.60} & \textbf{0.63} & \underline{0.44} & \underline{0.42} & \underline{0.43} & \textbf{6.56} & \textbf{6.37} & \textbf{6.47} \cr
USI3D$*$\dag \cite{liu2020cvpr}&  1.59 & 1.48 & 1.54 & 0.87 & 0.81 & 0.84 & 17.97 & 14.74 & 16.35 \cr
\midrule
Ours  ~& \textbf{0.57} & {0.71} & \underline{0.64} & \textbf{0.29} & \textbf{0.38} &  \textbf{0.34} & \underline{8.74} &  \underline{8.83} &\underline{8.79} \cr
\bottomrule[1.2pt]
\end{tabular}
\begin{tablenotes}
\footnotesize
        \item[$*$] Unsupervised methods but finetuned on the dataset. 
        \item[\dag] Using pre-trained parameters.  
\end{tablenotes}
\end{threeparttable}}
\caption{{Numerical results for MPI-Sintel }(image split). }
\vspace{-5mm}
\label{tab:sintel_image_split}.
\end{table}

\paragraph{\textbf{MPI-Sintel}} 
Unlike object-wise datasets, the MPI dataset provides color information for each pixel. To ensure a fair evaluation, we adopt the approach outlined in \cite{fan2018revisiting}, which employs scale-invariant MSE (si-MSE) and local scale-invariant MSE (si-LMSE).

We conduct a comparative analysis of our approach against several state-of-the-art methods (Table \ref{tab:sintel_image_split}). Our method surpasses others in terms of si-LMSE for albedo and shading, as well as si-MSE for albedo. Notably, methods like \cite{kim2016unified,barron2015shape,chen2013simple} also require RGB-D data to train their model. Furthermore, PoInt-Net demonstrates competitive performance for si-MSE shading and ranks second-best in DSSIM, highlighting its effectiveness in handling complex scenes with diverse lighting conditions. Notably, we do not compare with PIE-Net \cite{das2022pie}, due to it only reports results of scene split.
% Our method obtains 3-rd ranked result on si-MSE shading, can likely be attributed to the broad depth range of the MPI-Sintel dataset, which poses a challenge for PoInt-Net in separating lighting effects from depth variations. The improved performance on the local shading metric (si-LMSE) reflects this issue. 

The qualitative results, presented in Figure \ref{Fig:MPI-intrinsic}, highlight PoInt-Net's ability to generate high-quality results, particularly in terms of sharpness. This characteristic is advantageous for a point-based intrinsic network, where intrinsic features are processed on a point-by-point basis. In comparison to \cite{chen2013simple}, which also utilizes depth information for intrinsic decomposition, PoInt-Net consistently demonstrates improved reflectance and shading estimation. This improvement can be attributed mainly to the adoption of point cloud representation. For more results, please refer to the supplementary material.
\begin{figure*}[!t]
    % \centering
    \begin{minipage}{0.5\linewidth}
        \resizebox{0.75\linewidth}{!}{\includegraphics{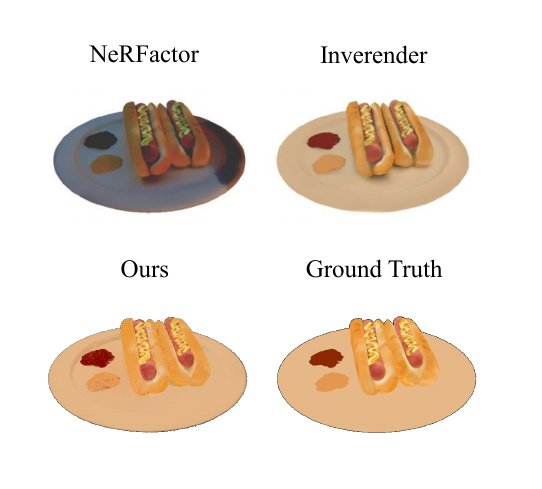}}
    \vspace{-2.8ex}
    \caption{Visual comparison with the NeRF based methods.}
    \label{Fig:All-NeRF}
    \end{minipage}
    \begin{minipage}{0.5\linewidth}
    \resizebox{\linewidth}{!}{\begin{tabular}{lc|c|c|c|c}
    \toprule
    Method & PSNR $\uparrow$ & SSIM $\uparrow$ & LPIPS $\downarrow$ & MSE $\downarrow$ & LMSE $\downarrow$ \\
    \hline
    NeRFactor\cite{zhang2021nerfactor} & 19.9167 & 0.9156 & 0.1354 & 0.0059 & 0.0210 \\
    PhySG\cite{zhang2021physg} & 23.3748 & 0.9231 & 0.1092 & 0.0034 & 0.0396 \\
    Inverender\cite{zhang2022modeling} & \underline{26.3078} & \underline{0.9380} & \underline{0.0572} & {0.0022} & 0.0226 \\
    Intrinsic-NeRF\cite{ye2023intrinsicnerf} & {24.2642} & {0.9371} & {0.0880} & \underline{0.0021} & {0.0173} \\
    \hline
    IIW\cite{bell2014intrinsic} & 22.0284 & 0.9307 & 0.0847 & 0.0099 & 0.0120 \\
    CGIntrinsic\cite{cgintrinsic} & 20.1583 & 0.9209 & 0.0996 & 0.0129 & 0.0141 \\
    USI3D\cite{liu2020cvpr} & 20.7571 & 0.9267 & 0.0887 & 0.0079 & 0.0149 \\
    Li \etal \cite{li2021openrooms}  & 16.8167 &0.8224   & 0.1661 & 0.0075&0.0089  \\
    PIE-Net\cite{das2022pie} & 18.8119 & 0.7870 & 0.2319 & 0.0163 & 0.0194 \\
    \hline
    Ours (zero-shot) & 22.4548 & 0.8986 & 0.1052 & {0.0059} &\underline{0.0071} \\
    Ours (finetuned)& \textbf{29.0404} & \textbf{0.9426} & \textbf{0.0543} & \textbf{0.0014} & \textbf{0.0015} \\
    \bottomrule
  \end{tabular}}
 \tabcaption{{Numerical results for Inverender dataset (reflectance)}.}
  \label{Tab: NeRF}
    \end{minipage}
\end{figure*}
\vspace{-1ex}
\paragraph{\textbf{Inverender}}
Recent advancements in NeRF methods have extended their functionalities to encompass scene intrinsic decomposition. In our evaluation, we contrast our approach with contemporary NeRF methodologies that similarly offer intrinsic decomposition capabilities (Table \ref{Tab: NeRF}).

By leveraging a pre-trained PoInt-Net, our approach effectively estimates reflectance for the Inverender dataset, achieving improved intrinsic image decomposition results comparable to NeRFactor \cite{zhang2021nerfactor}. This highlights our model's zero-shot estimation capability. Additionally, when fine-tuned on the Inverender dataset, PoInt-Net outperforms all competing methods, including NeRF-based techniques trained to over-fit a single scene (Figure \ref{Fig:All-NeRF}). For more visual comparisons, please refer to the supplemental.

\begin{figure}[t]
\begin{center}
\includegraphics[width=0.9\linewidth]{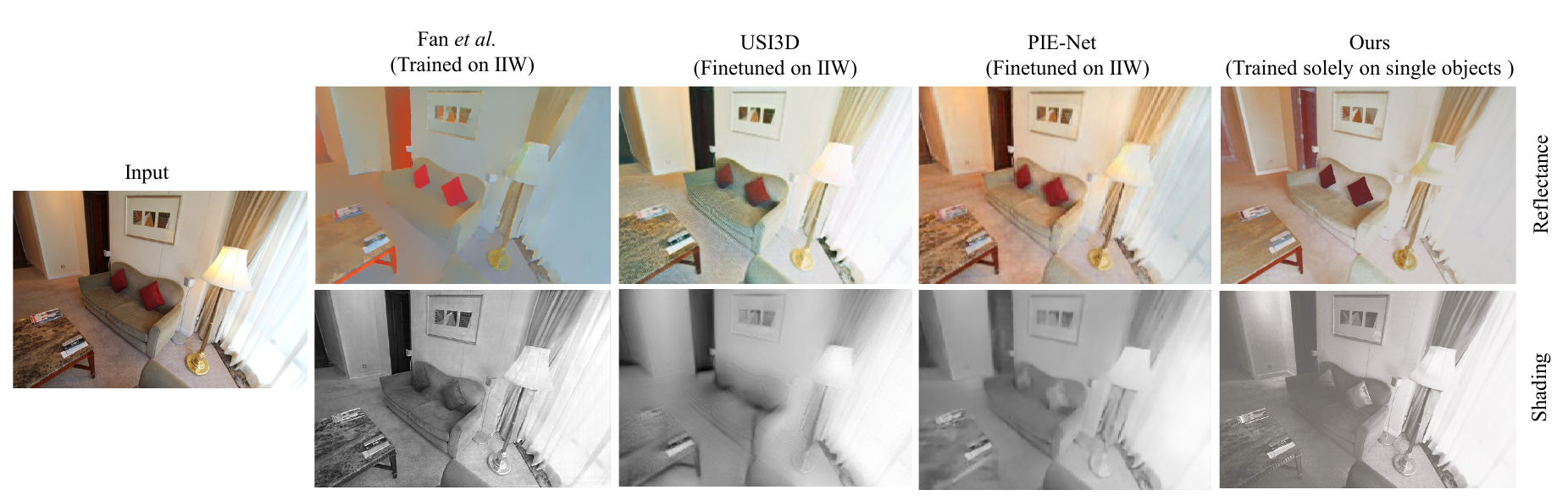}\\
% \captionsetup{font={footnotesize,stretch=1}, justification=raggedright}
\caption{Visual comparison on the IIW dataset \cite{bell2014intrinsic}. Our results are generated using the same model that we released for review (only trained on single-level objects). Results of \cite{fan2018revisiting} are taken from their paper (trained on IIW dataset with edge guidance).}
\label{fig:IIW}
\vspace{-2ex}
\end{center}
\end{figure}

\subsection{Real-world Generalization Performance}
\label{Sec:scene-result}
The real-world decomposition is conducted to further evaluate the generalization capability of our method.
\paragraph{\textbf{IIW dataset.}} Figure \ref{fig:IIW} provides visual results on the IIW dataset, using our model (only trained on a single object level dataset). The comparison is conducted with the visual results provided by \cite{fan2018revisiting} and estimated by \cite{liu2020cvpr}. Our method shows comparable results on the scene-level dataset without prior learning. Notably, our method effectively reconstructs the shape of objects, such as the red door in the background, which previous work \cite{fan2018revisiting} struggled to accomplish. 

The IIW dataset introduces WHDR as a performance metric, relying on human judgment scores to assess the accuracy of intrinsic image decompositions. Nevertheless, recent literature has raised doubts regarding the WHDR score's reliability in accurately gauging the performance of intrinsic decomposition methods \cite{forsyth2021intrinsic,nestmeyer2017reflectance,garces2022survey,careaga2023intrinsic}, e.g., \cite{nestmeyer2017reflectance} achieves a WHDR of 25.7\% by simply re-scaling the value of the original image into [0.55,1]. In line with \cite{careaga2023intrinsic}, we adjust our estimated reflectance by 0.5, yielding a comparable outcome, even when compared to models trained directly on the IIW dataset (Table \ref{Tab:IIW}).
\begin{table}[t]
\centering
  \resizebox{\linewidth}{!}{
    \begin{tabular}{lc|c|c|c|c|c|c|c}
    \toprule
     & Bell \etal\cite{bell2014intrinsic}    &  PIE-Net \cite{das2022pie}        & Careaga \etal\cite{careaga2023intrinsic} & USI3D \cite{liu2020cvpr}& Fan \etal\cite{fan2018revisiting} & IRISFormer\cite{zhu2022irisformer} &Ours & Ours$'$\\
    \hline
    WHDR(\%) &     20.6      & 21.3  & 24.9 &18.6&\underline{14.7}&\textbf{12.0}&26.7&{\textit{17.3}}\\
    Trained on IIW &   $\times$      &    $\times$   & $\times$  &\checkmark&\checkmark&\checkmark&$\times$&$\times$\\
    Type of the trained data &   S      &    S   & S  &S&S&S&O&O\\
    \bottomrule
    \end{tabular}}%
    \caption{{Results for the IIW dataset }\cite{bell2014intrinsic}. The letter 'S' and 'O' represent the scene-level and object-level datasets, respectively. Ours $'$ indicates the rescaled estimation, in line with \cite{careaga2023intrinsic}. }
  \label{Tab:IIW}%
\end{table}%

\begin{figure}[!t]
    \centering
    \resizebox{0.9\linewidth}{!}{\includegraphics{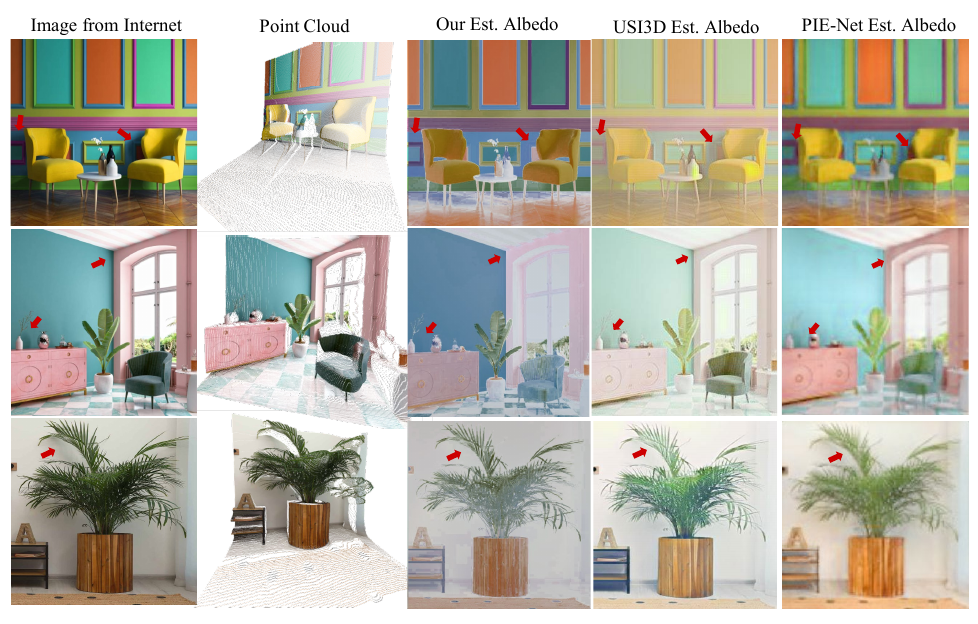}}
    \caption{Real-world intrinsics estimation comparison.  USI3D \cite{liu2020cvpr} and PIE-Net \cite{das2022pie} are trained or finetuned on large scene level datasets. Although our model is exclusively trained on datasets featuring single objects, our approach demonstrates the capability to accurately compute intrinsic properties for previously unseen objects and scenes. Notably, red arrows emphasize our exceptional estimation performance.}
    \label{Fig:Gen-Real}
\vspace{-2ex}
\end{figure}

Figure \ref{Fig:Gen-Real} demonstrates PoInt-Net's generalization to real-world images, randomly sourced from the internet. The point clouds are generated based on estimated depth maps obtained from \cite{Ranftl2022}. We employ the state-of-the-art unsupervised model \cite{liu2020cvpr} (finetuned on scene level datasets, e.g., CGI \cite{cgintrinsic}\& IIW \cite{bell2014intrinsic}) as a reference. Even though PoInt-Net is trained on a dataset focused on single objects (as described in Section \ref{Sec:Obj-result}), it shows the capability to accurately estimate surface reflectance and shading not only for single objects but also for complex scenes, as evidenced by its performance in regions such as the shadow area between the sofa and the wall. For further details and results, please refer to the supplementary materials.

% \subsection{Real-world Re-lighting/coloring}

\subsection{Point Cloud Matters}

 To demonstrate the effectiveness of point cloud representation in the task of intrinsic decomposition, we use deep CNNs (used in \cite{li2021openrooms}) and Vision Transformers (ViT, used in \cite{das2022pie,zhu2022irisformer}) as backbones and carefully design two alternative frameworks with a similar structure as our PoInt-Net. The same training strategy is applied, i.e, pre-training on the ShapeNet-Intrinsic dataset \cite{janner2017self} and fine-tuning on MIT-Intrinsic dataset \cite{grosse2009ground}. Table \ref{tab:RGBDvsPC} shows the point cloud representation superiority in intrinsic estimation with even fewer parameters. Interestingly, a few weeks before our paper submission, a parallel work in robotic learning \cite{zhu2024point} showcases a notable trend, "Point cloud-based methods, even those with the simplest designs, frequently surpass their RGB and RGB-D counterparts in performance." Although we are working on different subjects, a similar trend is also observed in our case. We argue that the community should not overlook the uniqueness of point cloud representation.

\begin{figure*}[!t]
\begin{minipage}{0.5\linewidth}
    \resizebox{0.85\linewidth}{!}{\begin{tabular}{lcc|c|l}
    \toprule
          & \multicolumn{2}{c}{MSE (10$^{-2}$)} & \multicolumn{1}{l}{Model}  & Data \\
    Backbone & \multicolumn{1}{l}{Albedo} & \multicolumn{1}{l|}{Shading} & \multicolumn{1}{l|}{Size (MB)}  & Format \\
    \midrule
    DeepCNN \cite{li2021openrooms} & 1.36  & 0.67  & $\sim$200 & \textit{RGB} \\
    ViT   \cite{das2022pie,zhu2022irisformer}&   1.15   & 0.66      &  $\sim$1450     & \textit{RGB} \\
    DeepCNN \cite{li2021openrooms} & 1.14  & 0.59  & $\sim$200    & \textit{RGB-D} \\
    ViT \cite{das2022pie,zhu2022irisformer}  &   1.58    &  1.06     & $\sim$1450         & \textit{RGB-D} \\
    \midrule
    Ours & \textbf{0.89} & \textbf{0.34} & \textbf{$\sim$20}  & PC \\
    \bottomrule
    \end{tabular}}%
    \tabcaption{{Evaluation of different data modalities} on MIT-Intrinsic dataset, all pre-trained on ShapeNet-Intrinsic dataset.}
  \label{tab:RGBDvsPC}%
\end{minipage}
\begin{minipage}{0.5\linewidth}
    \resizebox{0.85\linewidth}{!}{\begin{tabular}{l|cc|l}
    \toprule
          & \multicolumn{2}{c}{MSE (10$^{-2}$)}  & Data \\
    Method & \multicolumn{1}{l}{Albedo} & \multicolumn{1}{l|}{Shading}  & Format \\
    \midrule
    Janner \etal \cite{janner2017self}  &   3.03 \textcolor{green}{(0.30)$\downarrow$}  & 1.77 \textcolor{green}{(0.18)$\downarrow$}     & \textit{RGB-D} \\
    LapPyrNet \cite{cheng2018intrinsic}  & 1.72 \textcolor{green}{(0.05)$\downarrow$}     & 1.03 \textcolor{green}{(0.21)$\downarrow$}       & \textit{RGB-D} \\
    Fan \etal \cite{fan2018revisiting}   &   1.53 \textcolor{green}{(0.04)$\downarrow$}  &     0.81 \textcolor{green}{(0.07)$\downarrow$}  & \textit{RGB-D} \\
    % &      &          & \textit{RGB-D} \\
    \midrule
    Ours & \textbf{0.89} & \textbf{0.34} & PC \\
    \bottomrule
    \end{tabular}}%
    \tabcaption{{Evaluation of adding depth as extra input} on MIT-Intrinsic dataset, all pre-trained on ShapeNet-Intrinsic dataset. $(\cdot)\downarrow$ indicates error decreasing. }
  \label{tab:ExtraDepth}%
\end{minipage}
\end{figure*}
% \subsection{Point Clouds Matters}
\subsection{Ablation Study}

\paragraph{\textbf{Learnable shader.}} As discussed in Section \ref{Sec:Problem}, shading depends on the surface geometry. Table \ref{Tab:ShapeNet} illustrates the significant impact of the shader in improving shading quality numerically. Additionally, Figures \ref{Fig:SN-intrinsic} and \ref{Fig:MIT-intrinsic} demonstrate how the shader aids PoInt-Net in differentiating between invariant and ambient colors. This distinction is notably visible in areas such as the face and cloud segments of the "Sun" from MIT-Intrinsic, which have historically presented difficulties for numerous learning methods \cite{das2022pie, baslamisli2018joint, fan2018revisiting}. 
\paragraph{\textbf{Using depth information.}} We recognize the necessity of depth information in our methodology, which brings up issues regarding the equitable comparison with earlier intrinsic image decomposition (IID) techniques, especially given that only a select few incorporate depth and their implementations are not openly accessible. To ensure fairness in our assessment, we included depth information as an extra input in three publicly accessible IID frameworks \cite{fan2018revisiting, cheng2018intrinsic, janner2017self} (excluding PIE-Net \cite{das2022pie} due to its unavailability of training code). Table \ref{tab:ExtraDepth} demonstrates that, with depth information, performance of existing methods increased, yet not to the same level of performance as ours. This result further emphasizes the uniqueness and effectiveness of our proposed PoInt-Net.
\paragraph{\textbf{Depth quality.}} We assess the impact of depth quality in controlled conditions. PoIntNet demonstrates robust performance even with moderately noisy depth information. However, in highly noisy scenarios, there may be a decline in shading performance of up to 50\%, while albedo performance remains relatively stable, with a decrease of less than 10\%. These outcomes underscore the effectiveness of our method's design and are consistent with the fundamental principle of image formation, where shading is influenced by geometry. Please refer to supplemental for more details.
\paragraph{\textbf{Backbone selection.}} PoInt-Net processes the point cloud by using multiple MLP layers. We acknowledge that, there are many powerful point-based backbones, such as Point++\cite{qi2017pointnetplusplus}. We chose the relatively sampler backbone as it provides a good trade-off between computational efficiency and the ability to perform low-level vision tasks effectively, as evidenced in \cite{xing2022point}.
Please refer to supplemental for more details.

\begin{wrapfigure}{r}{7cm}
    \centering
    \resizebox{0.8\linewidth}{!}{\includegraphics{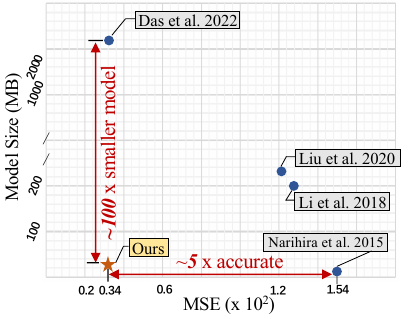}}
    \caption{Model size (MB) vs. MSE ($\times10^2$) on shading for the MIT-intrinsic dataset. Our method is very efficient, outperforming the state-of-the-art model with just 1/100 of the model size and achieving 5 times the accuracy of a model with a similar size.}
    \label{Fig:Size_MSE}
\end{wrapfigure}

\paragraph{\textbf{Model size of networks.}} As depicted in Figure \ref{Fig:Size_MSE}, our method excels in intrinsic estimation performance despite having a compact model size. The reported model size is in accordance with its official pre-trained model specifications. Overall, our approach maintains a smaller model size compared to others that incorporate additional components like a mapping module \cite{liu2020cvpr}, an adversarial network \cite{zoran2015learning}, a multi-scale CNN \cite{cheng2018intrinsic}, and a transformer \cite{das2022pie} in their network architecture.

\section{Conclusion}
We introduced point intrinsic representation and PoInt-Net for 3D-based intrinsic decomposition. PoInt-Net employs a point cloud representation to efficiently decompose surface light direction, reflectance, and shading maps, outperforming larger models on the MIT-intrinsic dataset while being highly efficient. Our experiments across different scenarios highlight its robustness and zero-shot generalization ability. Extra evaluation on the data modality evidenced point cloud as a valuable data format for intrinsic decomposition. 

\paragraph{\textbf{Limitations and Future Work.}}
Although PoInt-Net demonstrates stability with non-Lambertian scenes and multiple light sources, explicitly investigating such complex scenarios in the future is essential to extend the versatility of our approach. Additionally, our method works with a 3D point cloud derived from a 2D image, such conversion is not able to perform spatial relations like occlusion. Creating 3D dataset containing point clouds and intrinsic components will benefit the research in the future. Finally, some of our generalization results are only evaluated visually, due to the unavailability of either ground-truth or proper evaluation metrics. A quantitative evaluation process is needed in our future research.

\appendix
\section{Ablation Study}

\subsection{Depth quality}
Depth information plays a critical role in our method, particularly in shading estimation. 
To assess how our method performs under varying depth quality, we conducted a quantitative study focusing on two common types of depth inconsistencies:
\begin{compactitem}
    \item \textbf{Inaccurate Depth}: Here, we simulate depth inaccuracy by adding random values to a number of pixels. This introduces a type of noise that mimics the effect of depth imprecision.
    \item \textbf{Holes in Depth}: For this scenario, we randomly set a number of pixels to zero. This action creates 'holes' in the depth data, replicating instances where depth information is missing or unreliable.
\end{compactitem}
The number of pixels affected can be adjusted to various scales, simulating different levels of noise intensity. This allows for a more comprehensive evaluation of how our method is effected by depth inaccuracies ranging from minor to severe.

\begin{figure}[!t]
    \centering
    \resizebox{\linewidth}{!}{\includegraphics{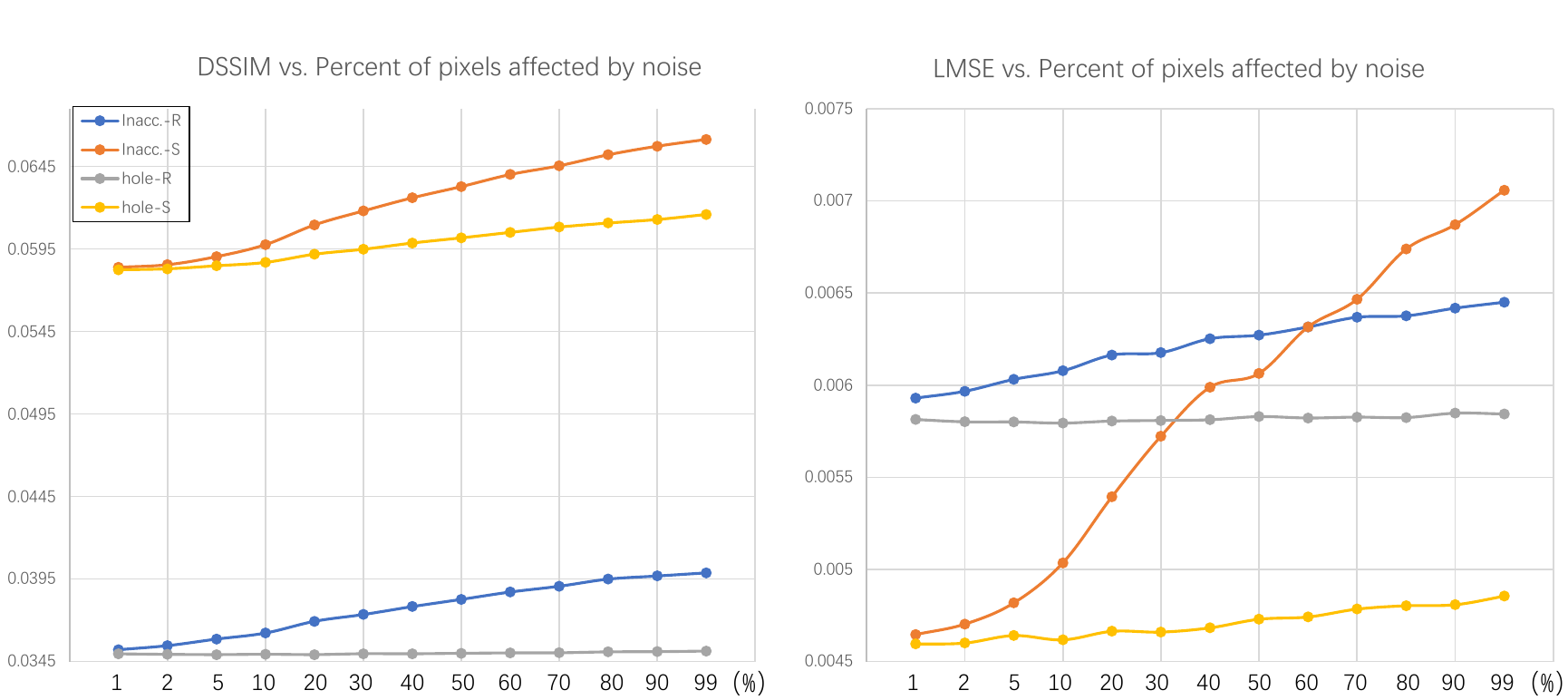}}
    \caption{DSSIM (left plot) and LMSE (right plot) on Shapenet-Intrinsic dataset vs. percent of pixels affected by noise. "Inacc." denotes inaccurate depth distribution, and "Hole" refers to the deletion in the depth. "R" represents reflectance, and "S" is for shading.}
    \label{Fig:noise effects}
    \vspace{-2ex}
\end{figure}

Figure \ref{Fig:noise effects} demonstrates albedo and shading performance under varying noise levels. Our method maintains stability with incomplete depth (e.g., holes). Yet, inaccurate depth significantly disrupts intrinsic estimation. Specifically, Shading estimation performance (LMSE) may decrease by 50\% under inaccurate depth.

\noindent{\textbf{Discussion}} Since our method is robust to noisy depth. The inclusion of depth data does not pose a considerable limitation to our method. Our approach requires only an RGB image alongside its estimated depth, providing adaptability across a diverse range of application scenarios. Please refer to Section \ref{Sec: More Result} to check our method's performance on complex scenes.
\subsection{Backbone selection}
Technically, several backbones can process RGB-D images. Table \ref{tab:advantage} summarizes their capabilities for IID. Our point cloud-based IID estimates real-world intrinsics with a smaller model size and shorter training time. Transformers and CNNs are not able to handle multiple size of images and are limited by the training time. NeRF based methods show good results yet lack of generalization ability. 
\begin{table}[t]
  \centering
  \resizebox{0.6\linewidth}{!}{
    \begin{tabular}{l|cccc}
    \toprule
          & \multicolumn{1}{c}{CNN} & \multicolumn{1}{c}{Transformer } &  \multicolumn{1}{c}{NeRF} & Ours\\
          &  \cite{baslamisli2021shadingnet,baslamisli2018cnn,fan2018revisiting,li2021openrooms}     &  \cite{das2022pie, zhu2022irisformer}    &\cite{zhang2021nerfactor,ye2023intrinsicnerf}  &\\
    \midrule
    Real-world generalize & some of them    &        \checkmark   & $\times$ &  \checkmark\\
    Model size&    $>$100MB       &$>$1GB   &  - &20MB\\
    Training Time &   $>$ 5 h     &    $>$ 1 d   & $>$ 1 d & $<$ 1.5 h\\
    Diverse input size &   $\times$     &    $\times$  & -  &\checkmark\\
    \end{tabular}}%
    \caption{Comparison with various IID methods using different backbones. }
  \label{tab:advantage}%
  \vspace{-4ex}
\end{table}%

\section{Details of Point Cloud Generation}

\paragraph{Building Point Cloud.} We generate the point cloud based on a pair of RGB image and its corresponding depth information. Normal information is pre-computed from the point cloud, which can be performed online or in advance for faster training and inference.

\paragraph{Camera Intrinsic Matrix.}
In Equation (4), the camera intrinsic matrix is employed to translate the point cloud from RGB-D image, i.e., images $\textbf{I}= [\textbf{I}_r,\textbf{I}_g,\textbf{I}_b] \in\mathbb{R}^{U\times{V}\times3}$ and depth map $\mathrm{D}\in\mathbb{R}^{U\times{V}\times1}$. For the MPI-Sintel dataset, we use the intrinsic matrix provide from the dataset. For other datasets, we use a default setting for the intrinsic matrix, where as the focal lengths and principal point are set as $(\frac{1}{2}{U},\frac{1}{2}{V})$, respectively. 
\section{Point Intrinsic Net: Architecture}
\begin{figure}[t]
    \centering
    \resizebox{0.8\linewidth}{!}{\includegraphics{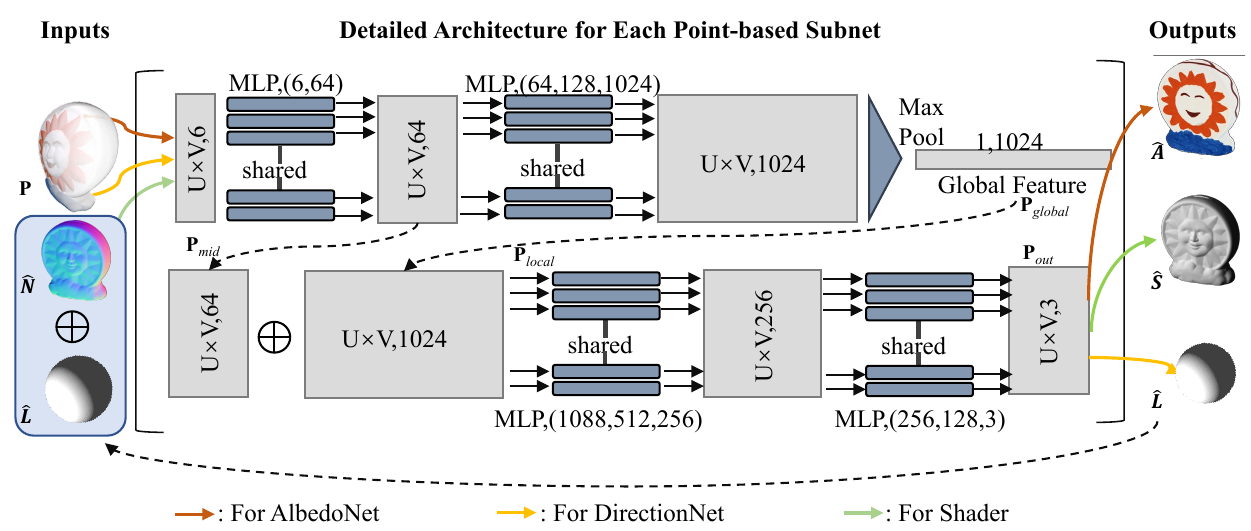}}
    \caption{The sub-net of PoInt-Net is based on a modified PointNet \cite{qi2017pointnet} architecture. For internal connections, refer to Figure 2 in the main paper. Three colors represent input/output for three subnets.}
    \label{Fig:SP_detail}
\end{figure}
Figure \ref{Fig:SP_detail} demonstrates the detailed architecture of the subnet of PoInt-Net. The input feature $\textbf{P}_{input}\in\mathbb{R}^{U\times{V}\times6}$ is first extracted by MLPs, and Max Pooling is employed to extract the significant global feature $\textbf{P}_{global}\in\mathbb{R}^{1\times1024}$. To maintain the ability in solving local information, the global feature is then repeated to the same number as the input points, and then concatenate with the inter-midden feature $\textbf{P}_{mid}\in\mathbb{R}^{U\times{V}\times{64}}$ to get the local feature $\textbf{P}_{local}\in\mathbb{R}^{U\times{V}\times1088}$. Finally, the output $\textbf{P}_{output}\in\mathbb{R}^{U\times{V}\times3}$ is obtained by MLPs which act as channel wise decoder. Since the three subnets share the similar network architecture, we use different color of the arrow to indicate the different input for the different nets, the color refer to the same color used in the Figure 2 of the main paper. 

\noindent{\textbf{Discussion}} Our choice of module configuration is based on empirical evidence:
1) \textit{Module Specialization.}: Given PointNet's limitations as anrelatively weaker encoder, we use three specialized modules for separate estimation of albedo, lighting, and shading.
2) \textit{Input Integration.} We found that concatenating surface normals with lighting, rather than multiplying them, yields improved outcomes. These insights will be added into the revision. 
% \subsection{}
\section{Cross Color Ratios Loss Calculation}
In paper, we propose the cross color ratios (CCR) loss to address the reflectance changes. Where the CCR $\{M_{RG},M_{RB},M_{GB}\}$ is calculated as:
\begin{equation}
    M_{RG} = \frac{R_{p1}G_{p2}}{R_{p2}G_{p1}}, M_{RB} = \frac{R_{p1}B_{p2}}{R_{p2}B_{p1}}, M_{GB} = \frac{G_{p1}B_{p2}}{G_{p2}B_{p1}},
\end{equation}
where the $\{R_{p1},G_{p1},B_{p1}\}$ and $\{R_{p2},G_{p2},B_{p2}\}$ represent the $R, G, B$ value at the two adjacent points $p_1$ and $p_2$, respectively.

\section{Implementation Details}

\paragraph{Environment Setting.}
PoInt-Net is developed using the PyTorch framework, enabling it to run on both CPU and GPU platforms. During evaluation, our method runs on a single NVIDIA 1080Ti GPU, with inference times dependent on the image size. For a 512$\times$512 image, the average inference time is $\sim$ 10 frames per second. It's important to note that this inference time is achieved without any pruning techniques applied.
\paragraph{Training Details.} For single object datasets, the point cloud is sampled by voxel downsampling where the voxel size is set to 0.03. For scene datasets, the point cloud is resized to $64\times64$ points by average downsampling. The batch size is set based on the GPU memory accordingly. Adam \cite{kingma2014adam} is employed as the optimizer. The learning rate is $3\times10^{-4}$. All networks are trained till convergences.

\section{More Results: Intrinsic Decomposition}
\label{Sec: More Result}
\begin{figure*}[!t]
    \centering
    \resizebox{0.9\textwidth}{!}{\includegraphics{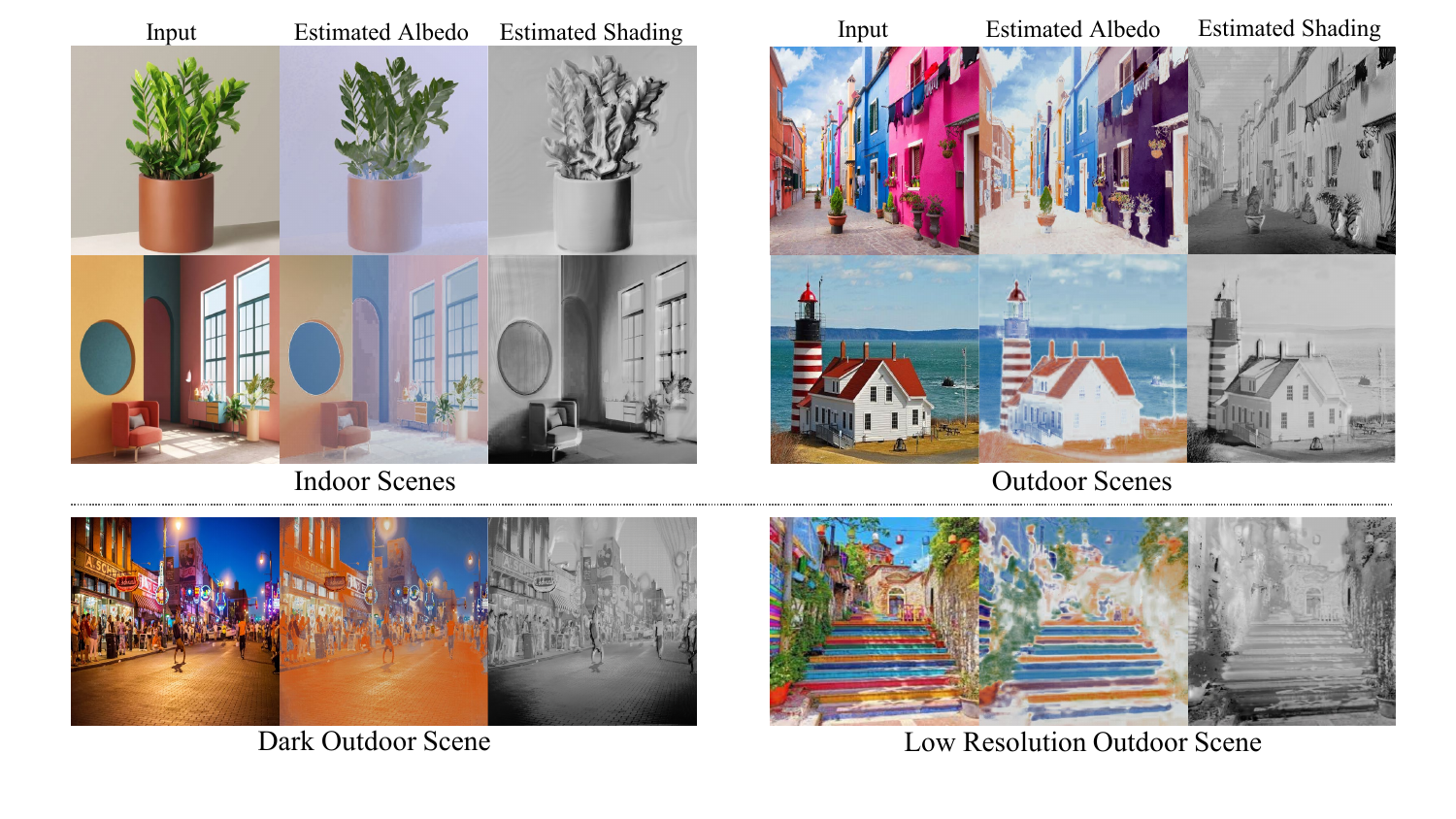}}
    \caption{Visualization results across different shapes, scenes, and domains in the real world. Our method consistently estimates the reasonable results. Note: our method is only trained on single-object level datasets.}
    \label{Fig:All-Real}
\end{figure*}
\paragraph{Real World.}

The use of monocular depth estimation in PoInt-Net allows for effective decomposition of intrinsic properties in real-world scenes. Figure \ref{Fig:All-Real} showcases additional visual results across various environments, both indoor and outdoor. PoInt-Net consistently delivers reliable reflectance and shading estimates. Notably, it performs well even in challenging conditions like dark or low-resolution outdoor scenes, accurately decomposing invariant colors and shading. 

As discussed in the paper, PoInt-Net is trained on single-object level datasets, i.e., ShapeNet-Intrinsic and MIT-Intrinsic. However, our method still achieves impressive intrinsic decomposition results across various shapes, scenes, and domains. These findings suggest that PoInt-Net is capable of solving intrinsic decomposition tasks at a very low-level, highlighting its robustness and versatility.

\begin{figure*}[!t]
    \centering
    \resizebox{0.9\textwidth}{!}{\includegraphics{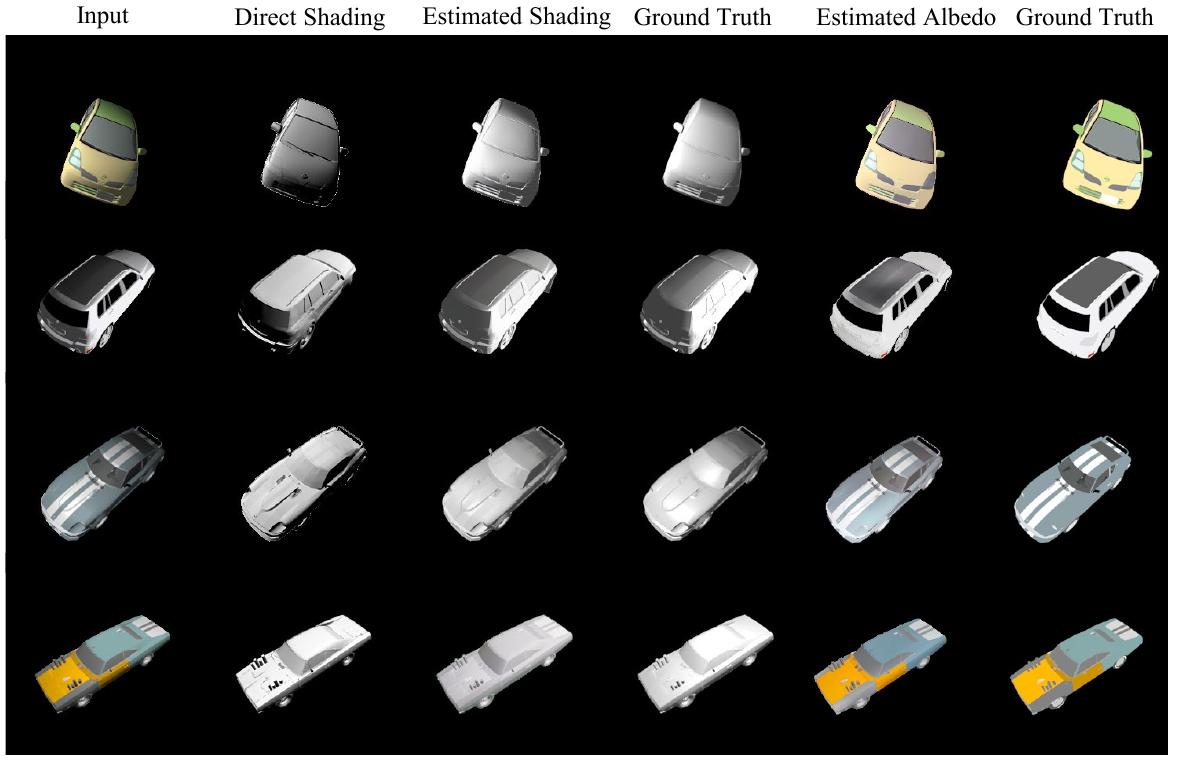}}
    \caption{Qualitative results on ShapeNet-Intrinsic dataset. The Direct shading is computed by surface normal and surface light direction.}
    \label{Fig:All-SP}
\end{figure*}

\begin{figure*}[!t]
    \centering
    \resizebox{0.9\textwidth}{!}{\includegraphics{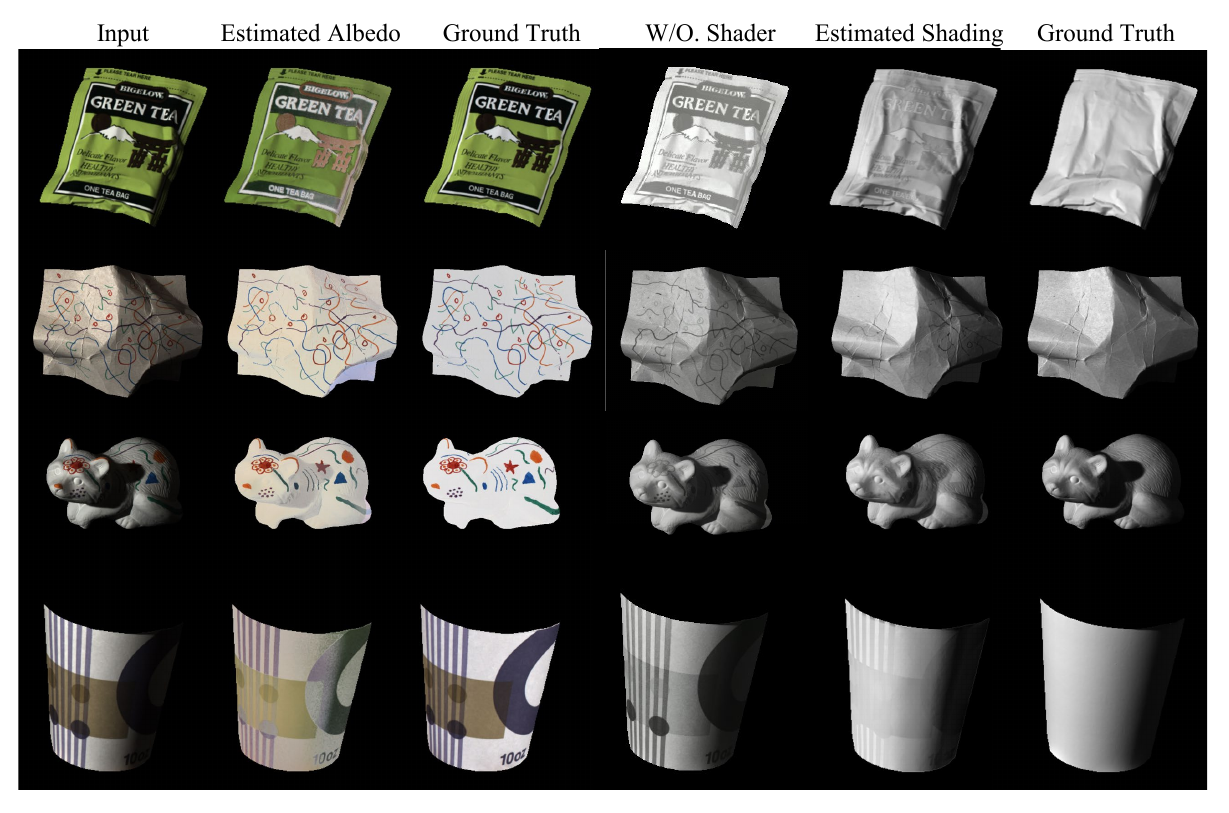}}
    \caption{Qualitative results on MIT-Intrinsic dataset. The ablation study on the shader (W/O. Shader) is applied. }
    \label{Fig:All-MIT}
\end{figure*}

\begin{figure*}[!t]
    \centering
    \resizebox{0.9\textwidth}{!}{\includegraphics[angle=-90]{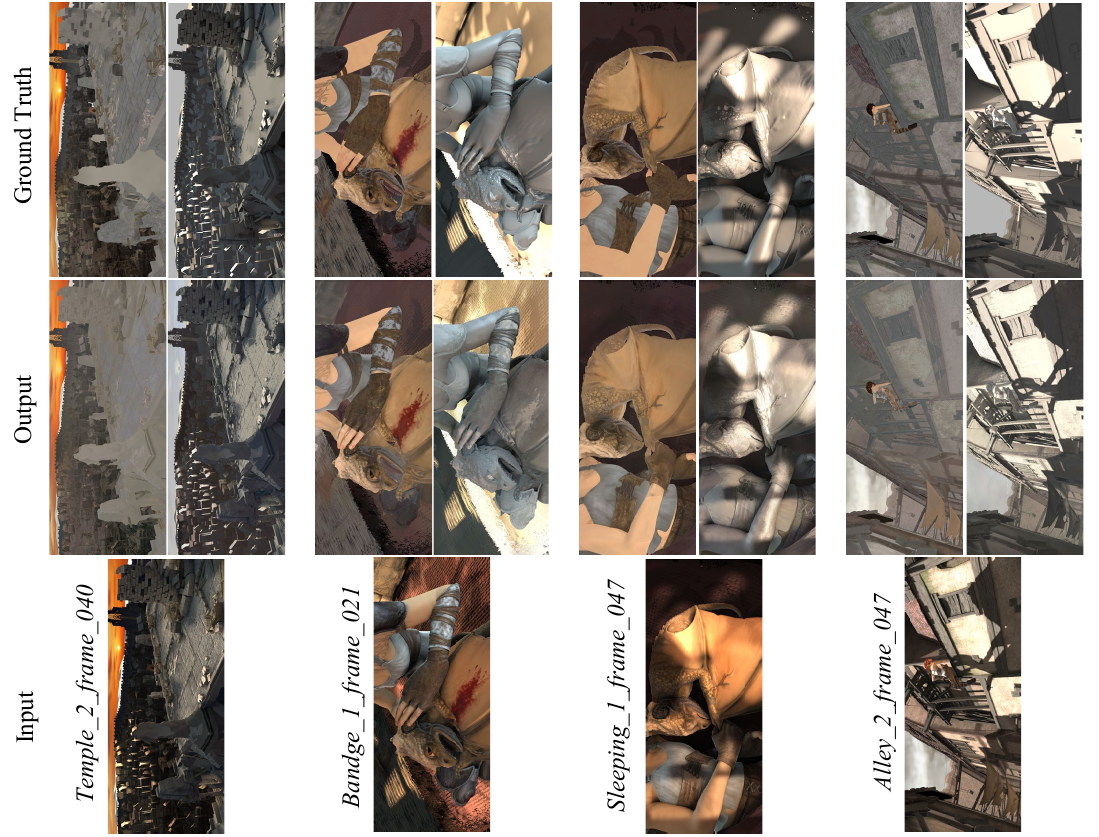}}
    \caption{Visual results on MPI-Sintel dataset.}
    \label{Fig:All-MPI}
\end{figure*}

\begin{figure}[!t]
    \centering
    \resizebox{0.5\linewidth}{!}{\includegraphics{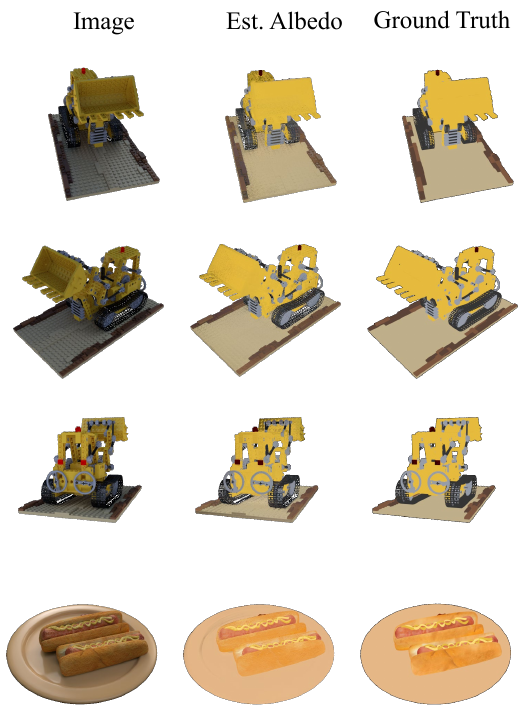}}
    \caption{Qualitative results on Inverender dataset. }
    \label{Fig:NeRF-More}
\end{figure}

\begin{figure}[!t]
    \centering
    \resizebox{0.5\linewidth}{!}{\includegraphics{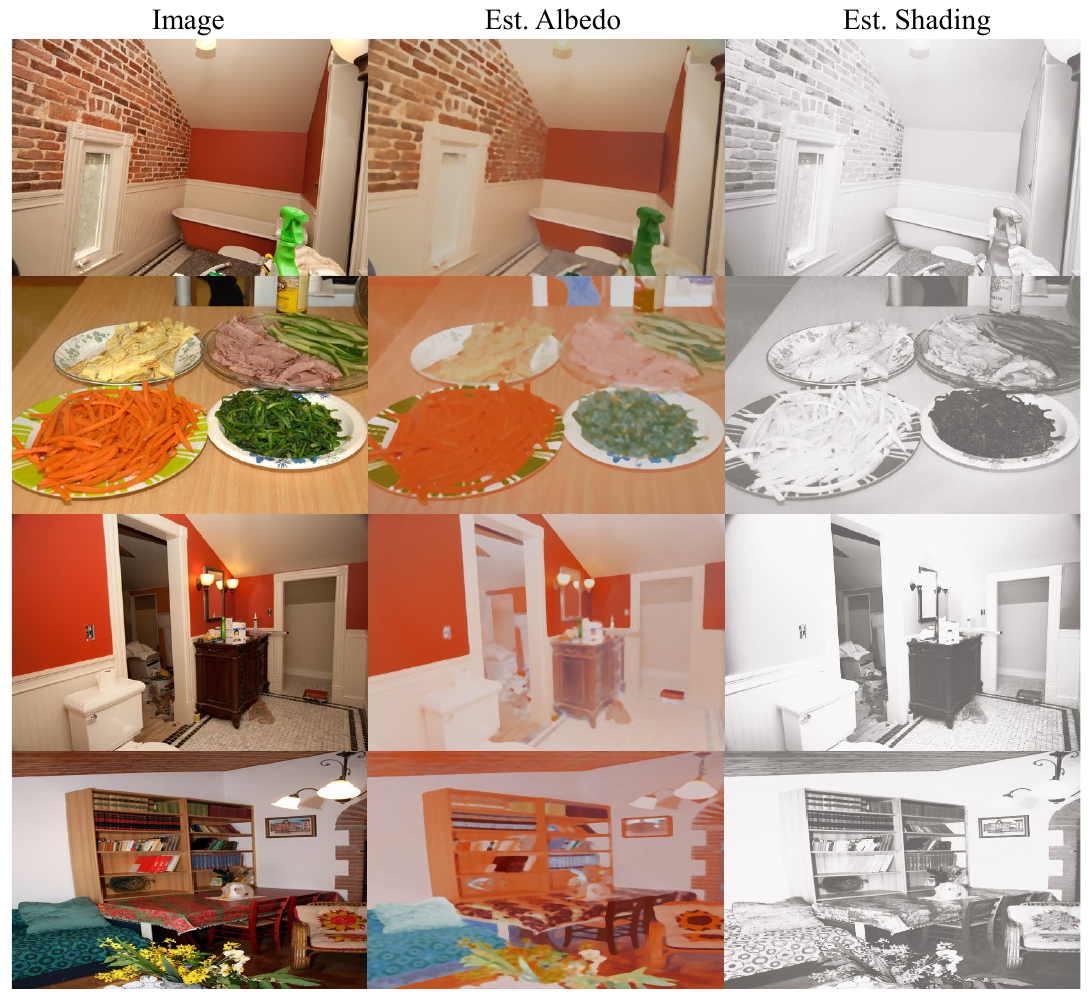}}
    \caption{Qualitative results on IIW dataset. Note: Our method is not trained on IIW dataset.}
    \label{Fig:IIW-More}
\end{figure}

\paragraph{ShapeNet-Intrinsic.}
We present additional results on the ShapeNet dataset in Figure \ref{Fig:All-SP}. These results showcase PoInt-Net's ability to estimate surface light direction, with direct shading results displayed in the second column of the figure.
\paragraph{MIT-Intrinsic.}
Figure \ref{Fig:All-MIT} presents additional results on the MIT-Intrinsic dataset, including ablation study results for the shader. The findings demonstrate that the shader significantly enhances PoInt-Net's ability to distinguish between invariant color and illumination.
\paragraph{MPI-Sintel.}
Figure \ref{Fig:All-MPI} presents extensional results that provide evidence for PoInt-Net's ability to handle complex scenes during intrinsic decomposition training. The findings suggest that the architecture of PoInt-Net is capable of undertaking such tasks effectively.
\paragraph{Inverender.}
Figure \ref{Fig:NeRF-More} shows the additional results of the reflectance estimation on the Inverender dataset.
\paragraph{IIW-Dataset.}
Figure \ref{Fig:IIW-More} provides more visual results of the intrinsic estimation on IIW dataset \cite{bell2014intrinsic}. 
\clearpage
\bibliographystyle{splncs04}
\bibliography{main}

\begin{thebibliography}{10}
\providecommand{\url}[1]{\texttt{#1}}
\providecommand{\urlprefix}{URL }
\providecommand{\doi}[1]{https://doi.org/#1}

\bibitem{barron2015shape}
Barron, J.T., Malik, J.: Shape, illumination, and reflectance from shading. IEEE TPAMI  \textbf{37}(8),  1670--1687 (2015)

\bibitem{barrow1978recovering}
Barrow, H., Tenenbaum, J., Hanson, A., Riseman, E.: Recovering intrinsic scene characteristics. Comput. Vis. Syst  \textbf{2},  3--26 (1978)

\bibitem{baslamisli2021shadingnet}
Baslamisli, A.S., Das, P., Le, H.A., Karaoglu, S., Gevers, T.: Shadingnet: Image intrinsics by fine-grained shading decomposition. IJCV  \textbf{129}(8),  2445--2473 (2021)

\bibitem{baslamisli2018joint}
Baslamisli, A.S., Groenestege, T.T., Das, P., Le, H.A., Karaoglu, S., Gevers, T.: Joint learning of intrinsic images and semantic segmentation. In: ECCV (2018)

\bibitem{baslamisli2018cnn}
Baslamisli, A.S., Le, H.A., Gevers, T.: Cnn based learning using reflection and retinex models for intrinsic image decomposition. In: CVPR (2018)

\bibitem{bell2014intrinsic}
Bell, S., Bala, K., Snavely, N.: Intrinsic images in the wild. ACM Transactions on Graphics (TOG)  \textbf{33}(4), ~159 (2014)

\bibitem{buchsbaum1980spatial}
Buchsbaum, G.: A spatial processor model for object colour perception. Journal of the Franklin Institute  \textbf{310}(1),  1--26 (1980)

\bibitem{butler2012naturalistic}
Butler, D.J., Wulff, J., Stanley, G.B., Black, M.J.: A naturalistic open source movie for optical flow evaluation. In: ECCV (2012)

\bibitem{careaga2023intrinsic}
Careaga, C., Aksoy, Y.: Intrinsic image decomposition via ordinal shading. ACM Transactions on Graphics  (2023)

\bibitem{shapenet2015}
Chang, A.X., Funkhouser, T., Guibas, L., Hanrahan, P., Huang, Q., Li, Z., Savarese, S., Savva, M., Song, S., Su, H., Xiao, J., Yi, L., Yu, F.: {ShapeNet: An Information-Rich 3D Model Repository}. Tech. Rep. arXiv:1512.03012 [cs.GR], Stanford University --- Princeton University --- Toyota Technological Institute at Chicago (2015)

\bibitem{chen2013simple}
Chen, Q., Koltun, V.: A simple model for intrinsic image decomposition with depth cues. In: ICCV (2013)

\bibitem{cheng2018intrinsic}
Cheng, L., Zhang, C., Liao, Z.: Intrinsic image transformation via scale space decomposition. In: CVPR (2018)

\bibitem{das2022pie}
Das, P., Karaoglu, S., Gevers, T.: Pie-net: Photometric invariant edge guided network for intrinsic image decomposition. In: CVPR (2022)

\bibitem{fan2018revisiting}
Fan, Q., Yang, J., Hua, G., Chen, B., Wipf, D.: Revisiting deep intrinsic image decompositions. In: CVPR (2018)

\bibitem{forsyth2021intrinsic}
Forsyth, D., Rock, J.J.: Intrinsic image decomposition using paradigms. IEEE TPAMI  \textbf{44}(11),  7624--7637 (2021)

\bibitem{garces2022survey}
Garces, E., Rodriguez-Pardo, C., Casas, D., Lopez-Moreno, J.: A survey on intrinsic images: Delving deep into lambert and beyond. IJCV  \textbf{130}(3),  836--868 (2022)

\bibitem{gevers1999color}
Gevers, T., Smeulders, A.W.: Color-based object recognition. Pattern recognition  \textbf{32}(3),  453--464 (1999)

\bibitem{grosse2009ground}
Grosse, R., Johnson, M.K., Adelson, E.H., Freeman, W.T.: Ground truth dataset and baseline evaluations for intrinsic image algorithms. In: ICCV (2009)

\bibitem{hachama2015intrinsic}
Hachama, M., Ghanem, B., Wonka, P.: Intrinsic scene decomposition from rgb-d images. In: Proceedings of the IEEE International Conference on Computer Vision. pp. 810--818 (2015)

\bibitem{janner2017self}
Janner, M., Wu, J., Kulkarni, T.D., Yildirim, I., Tenenbaum, J.: Self-supervised intrinsic image decomposition. In: NIPS (2017)

\bibitem{kajiya1986rendering}
Kajiya, J.T.: The rendering equation. In: SIGGRAPH. pp. 143--150 (1986)

\bibitem{kim2016unified}
Kim, S., Park, K., Sohn, K., Lin, S.: Unified depth prediction and intrinsic image decomposition from a single image via joint convolutional neural fields. In: ECCV (2016)

\bibitem{kingma2014adam}
Kingma, D.P., Ba, J.: Adam: A method for stochastic optimization. arXiv preprint arXiv:1412.6980  (2014)

\bibitem{krizhevsky2012imagenet}
Krizhevsky, A., Sutskever, I., Hinton, G.E.: Imagenet classification with deep convolutional neural networks. In: NIPS (2012)

\bibitem{laffont2012coherent}
Laffont, P.Y., Bousseau, A., Paris, S., Durand, F., Drettakis, G.: Coherent intrinsic images from photo collections. ACM ToG  \textbf{31}(6) (2012)

\bibitem{lee2012estimation}
Lee, K.J., Zhao, Q., Tong, X., Gong, M., Izadi, S., Lee, S.U., Tan, P., Lin, S.: Estimation of intrinsic image sequences from image+ depth video. In: ECCV (2012)

\bibitem{lettry2018darn}
Lettry, L., Vanhoey, K., Van~Gool, L.: Darn: a deep adversarial residual network for intrinsic image decomposition. In: WACV (2018)

\bibitem{cgintrinsic}
Li, Z., Snavely, N.: Cgintrinsics: Better intrinsic image decomposition through physically-based rendering. In: ECCV (2018)

\bibitem{li2018learning}
Li, Z., Snavely, N.: Learning intrinsic image decomposition from watching the world. CVPR  (2018)

\bibitem{li2021openrooms}
Li, Z., Yu, T.W., Sang, S., Wang, S., Song, M., Liu, Y., Yeh, Y.Y., Zhu, R., Gundavarapu, N., Shi, J., et~al.: Openrooms: An open framework for photorealistic indoor scene datasets. In: CVPR. pp. 7190--7199 (2021)

\bibitem{liu2020cvpr}
Liu, Y., Li, Y., You, S., Lu, F.: Unsupervised learning for intrinsic image decomposition from a single image. In: CVPR (2020)

\bibitem{luo2020niid}
Luo, J., Huang, Z., Li, Y., Zhou, X., Zhang, G., Bao, H.: Niid-net: adapting surface normal knowledge for intrinsic image decomposition in indoor scenes. IEEE TVCG  \textbf{26}(12),  3434--3445 (2020)

\bibitem{ma2018single}
Ma, W.C., Chu, H., Zhou, B., Urtasun, R., Torralba, A.: Single image intrinsic decomposition without a single intrinsic image. In: ECCV (2018)

\bibitem{mildenhall2020nerf}
Mildenhall, B., Srinivasan, P.P., Tancik, M., Barron, J.T., Ramamoorthi, R., Ng, R.: Nerf: Representing scenes as neural radiance fields for view synthesis. In: ECCV. pp. 405--421. Springer (2020)

\bibitem{narihira2015direct}
Narihira, T., Maire, M., Yu, S.X.: Direct intrinsics: Learning albedo-shading decomposition by convolutional regression. In: ICCV (2015)

\bibitem{nestmeyer2017reflectance}
Nestmeyer, T., Gehler, P.V.: Reflectance adaptive filtering improves intrinsic image estimation. In: CVPR (2017)

\bibitem{nicodemus1965directional}
Nicodemus, F.E.: Directional reflectance and emissivity of an opaque surface. Applied optics  \textbf{4}(7),  767--775 (1965)

\bibitem{qi2017pointnet}
Qi, C.R., Su, H., Mo, K., Guibas, L.J.: Pointnet: Deep learning on point sets for 3d classification and segmentation. In: CVPR. pp. 652--660 (2017)

\bibitem{qi2017pointnetplusplus}
Qi, C.R., Yi, L., Su, H., Guibas, L.J.: Pointnet++: Deep hierarchical feature learning on point sets in a metric space. arXiv preprint arXiv:1706.02413  (2017)

\bibitem{qian2021fast}
Qian, Y., Shi, M., Kamarainen, J.K., Matas, J.: Fast fourier intrinsic network. In: WACV (2021)

\bibitem{Ranftl2022}
Ranftl, R., Lasinger, K., Hafner, D., Schindler, K., Koltun, V.: Towards robust monocular depth estimation: Mixing datasets for zero-shot cross-dataset transfer. IEEE TPAMI  \textbf{44}(3) (2022)

\bibitem{rother2011recovering}
Rother, C., Kiefel, M., Zhang, L., Sch{\"o}lkopf, B., Gehler, P.V.: Recovering intrinsic images with a global sparsity prior on reflectance. In: NIPS (2011)

\bibitem{sato2023unsupervised}
Sato, S., Yao, Y., Yoshida, T., Kaneko, T., Ando, S., Shimamura, J.: Unsupervised intrinsic image decomposition with lidar intensity. In: CVPR. pp. 13466--13475 (2023)

\bibitem{sengupta2018sfsnet}
Sengupta, S., Kanazawa, A., Castillo, C.D., Jacobs, D.W.: Sfsnet: Learning shape, reflectance and illuminance of facesin the wild'. In: CVPR. pp. 6296--6305 (2018)

\bibitem{shen2011intrinsic}
Shen, L., Yeo, C.: Intrinsic images decomposition using a local and global sparse representation of reflectance  (2011)

\bibitem{shen2013intrinsic}
Shen, L., Yeo, C., Hua, B.S.: Intrinsic image decomposition using a sparse representation of reflectance. IEEE TPAMI  \textbf{35}(12),  2904--2915 (2013)

\bibitem{shi2017learning}
Shi, J., Dong, Y., Su, H., Stella, X.Y.: Learning non-lambertian object intrinsics across shapenet categories. In: CVPR (2017)

\bibitem{xing2022point}
Xing, X., Qian, Y., Feng, S., Dong, Y., Matas, J.: Point cloud color constancy. In: CVPR (2022)

\bibitem{xu2022point}
Xu, Q., Xu, Z., Philip, J., Bi, S., Shu, Z., Sunkavalli, K., Neumann, U.: Point-nerf: Point-based neural radiance fields. In: CVPR (2022)

\bibitem{ye2023intrinsicnerf}
Ye, W., Chen, S., Bao, C., Bao, H., Pollefeys, M., Cui, Z., Zhang, G.: Intrinsicnerf: Learning intrinsic neural radiance fields for editable novel view synthesis. In: ICCV. pp. 339--351 (2023)

\bibitem{zhang2021physg}
Zhang, K., Luan, F., Wang, Q., Bala, K., Snavely, N.: Physg: Inverse rendering with spherical gaussians for physics-based material editing and relighting. In: CVPR. pp. 5453--5462 (2021)

\bibitem{zhang2021nerfactor}
Zhang, X., Srinivasan, P.P., Deng, B., Debevec, P., Freeman, W.T., Barron, J.T.: Nerfactor: Neural factorization of shape and reflectance under an unknown illumination. ACM TOG  \textbf{40}(6),  1--18 (2021)

\bibitem{zhang2022modeling}
Zhang, Y., Sun, J., He, X., Fu, H., Jia, R., Zhou, X.: Modeling indirect illumination for inverse rendering. In: CVPR. pp. 18643--18652 (2022)

\bibitem{zhou2015learning}
Zhou, T., Krahenbuhl, P., Efros, A.A.: Learning data-driven reflectance priors for intrinsic image decomposition. In: CVPR (2015)

\bibitem{zhu2024point}
Zhu, H., Wang, Y., Huang, D., Ye, W., Ouyang, W., He, T.: Point cloud matters: Rethinking the impact of different observation spaces on robot learning. arXiv preprint arXiv:2402.02500  (2024)

\bibitem{zhu2022irisformer}
Zhu, R., Li, Z., Matai, J., Porikli, F., Chandraker, M.: Irisformer: Dense vision transformers for single-image inverse rendering in indoor scenes. In: CVPR. pp. 2822--2831 (2022)

\bibitem{zoran2015learning}
Zoran, D., Isola, P., Krishnan, D., Freeman, W.T.: Learning ordinal relationships for mid-level vision. In: CVPR (2015)

\end{thebibliography}
\end{document}